\documentclass[journal]{IEEEtran}
\usepackage{amsmath}
\usepackage{amsthm}

\usepackage{cancel}

\usepackage{amssymb}
\usepackage{amsfonts}
\usepackage{utfsym} 
\usepackage[colorlinks=true,urlcolor=black]{hyperref}
\usepackage{cleveref}
\crefname{section}{Sec.}{Secs.}
\Crefname{section}{Section}{Sections}
\Crefname{table}{Table}{Tables}
\crefname{table}{Tab.}{Tabs.}
\Crefname{figure}{Figure}{Figures}
\crefname{figure}{Fig.}{Figs.}
\Crefname{equation}{Equation}{Equations}
\crefname{equation}{Eq.}{Eqs.}
\crefname{algorithm}{Alg.}{Algs.}
\Crefname{algorithm}{Algorithm}{Algorithms}
\usepackage{url} 
\usepackage{cite} 
\usepackage{graphicx}
\usepackage{multirow}
\usepackage{makecell}
\usepackage[flushleft]{threeparttable}
\usepackage{colortbl}
\usepackage{booktabs}
\usepackage{caption}
\usepackage{subcaption}
\usepackage{xspace}
\usepackage{xcolor}
\usepackage{enumitem}
\usepackage[most]{tcolorbox} 
\usepackage{algorithm}
\usepackage{algpseudocode} 

\def\eg{\emph{e.g.,}\xspace}

\def\ie{\emph{i.e.,}\xspace}

\def\vs{\emph{vs.}\xspace}

\newcommand{\one}{({\em i})\xspace}
\newcommand{\two}{({\em ii})\xspace}
\newcommand{\three}{({\em iii})\xspace}
\newcommand{\four}{({\em iv})\xspace}

\tcbset{
  takeaway/.style={
    colback=gray!10,       
    colframe=gray!60,      
    boxrule=0.8pt,         
    arc=4mm,               
    left=5mm, right=5mm,   
    top=3mm, bottom=3mm,   
    fonttitle=\bfseries,
    coltitle=black
  }
}

\ifCLASSINFOpdf
\else
\fi
\begin{document}
%
\title{GreedyPixel: Fine-Grained Black-Box\\Adversarial Attack Via Greedy Algorithm}
%
%
%

\author{Hanrui Wang$^*$,~
        Ching-Chun Chang,~
        Chun-Shien Lu,~
        Christopher Leckie,~
        and Isao Echizen,~
\thanks{$^*$Corresponding author.}
\thanks{Hanrui Wang, Ching-Chun Chang, and Isao Echizen are with Echizen Lab, National Institute of Informatics (NII), Tokyo, Japan, e-mail: \{hanrui\_wang, ccchang, iechizen\}@nii.ac.jp. Chun-Shien Lu is with Institute of Information Science, Academia Sinica, Taipei, Taiwan, e-mail: lcs@iis.sinica.edu.tw. Christopher Leckie is with the School of Computing and Information Systems, the University of Melbourne, Australia, e-mail: caleckie@unimelb.edu.au.}
\thanks{This work was partially supported by JSPS KAKENHI Grants JP21H04907 and JP24H00732, by JST CREST Grant JPMJCR20D3 and JPMJCR2562 including AIP challenge program, by JST AIP Acceleration Grant JPMJCR24U3, and by JST K Program Grant JPMJKP24C2 Japan.}
}

%
%

\markboth{IEEE Transactions on Information Forensics and Security,~Vol.~20, November~2025}%
{Wang \MakeLowercase{\textit{et al.}}: GreedyPixel: Fine-Grained Black-Box Adversarial Attack Via Greedy Algorithm}
%



\maketitle

\begin{abstract}
Deep neural networks are highly vulnerable to adversarial examples, which are inputs with small, carefully crafted perturbations that cause misclassification—making adversarial attacks a critical tool for evaluating robustness. Existing black-box methods typically entail a trade-off between precision and flexibility: pixel-sparse attacks (e.g., single- or few-pixel attacks) provide fine-grained control but lack adaptability, whereas patch- or frequency-based attacks improve efficiency or transferability, but at the cost of producing larger and less precise perturbations. We present \emph{GreedyPixel}, a fine-grained black-box attack method that performs \emph{brute-force-style, per-pixel greedy optimization} guided by a surrogate-derived priority map and refined by means of query feedback. It evaluates each coordinate directly \emph{without any gradient information}, guaranteeing monotonic loss reduction and convergence to a coordinate-wise optimum, while also yielding near white-box-level precision and pixel-wise sparsity and perceptual quality. On the CIFAR-10 and ImageNet datasets, spanning convolutional neural networks (CNNs) and Transformer models, GreedyPixel achieved state-of-the-art success rates with visually imperceptible perturbations, effectively bridging the gap between black-box practicality and white-box performance. The implementation is available at \url{https://github.com/azrealwang/greedypixel}.
\end{abstract}

\begin{IEEEkeywords}
Black-box adversarial attack, greedy algorithm.
\end{IEEEkeywords}

%
\IEEEpeerreviewmaketitle

\section{Introduction}
Deep neural networks have achieved state-of-the-art performance in vision, language, and multimodal tasks, yet they remain vulnerable to adversarial examples, which are inputs with small, carefully crafted perturbations that cause incorrect predictions. Adversarial attacks are therefore a critical tool for assessing model robustness, particularly in safety-critical applications such as face recognition \cite{deng2019arcface}, medical diagnosis \cite{liu2019comparison}, and multimodal reasoning \cite{radford2021learning}. Existing adversarial attacks can be characterized along four complementary dimensions, which together reveal the gap that this paper investigates.

\begin{table*}[!t]
    \centering
    \caption{Related adversarial attacks and their characteristics.}
    \label{tab_rw}
    \begin{threeparttable}
        \setlength{\tabcolsep}{2.3mm}{\begin{tabular}{l|c|c|c|c|c|c}
            \toprule
            Attack&Strategy&White-box&Black-box&High ASR&Minimal Changes&Direct Optimization\\
            \midrule
            FGSM~\cite{goodfellow2014explaining} [2014]&Gradient-based&$\usym{2713}$&$\usym{2717}$&$\usym{2713}$&$\usym{2717}$&$\usym{2713}$\\
            CW~\cite{carlini2017towards} [2017]&Gradient-based&$\usym{2713}$&$\usym{2717}$&$\usym{2713}$&$\usym{2713}$&$\usym{2713}$\\
            PGD~\cite{madry2018towards} [2018]&Gradient-based&$\usym{2713}$&$\usym{2717}$&$\usym{2713}$&$\usym{2717}$&$\usym{2713}$\\
            GreedyFool~\cite{dong2020greedyfool} [2020]&Gradient-based&$\usym{2713}$&$\usym{2717}$&$\usym{2713}$&$\usym{2713}$&$\usym{2713}$\\
            APGD~\cite{croce2020reliable} [2020]$^*$&Gradient-based&$\usym{2713}$&$\usym{2717}$&$\usym{2713}$&$\usym{2717}$&$\usym{2713}$\\
            AutoAttack~\cite{croce2020reliable} [2020]$^*$&Ensembling white and black&$\usym{2713}$&$\usym{2717}$&$\usym{2713}$&$\usym{2717}$&$\usym{2713}$\\
            PBA~\cite{moon2019parsimonious} [2019]&Submodular searching&$\usym{2717}$&$\usym{2713}$&$\usym{2713}$&$\usym{2717}$&$\usym{2713}$\\
            Square Attack~\cite{andriushchenko2020square} [2020]&Submodular searching&$\usym{2717}$&$\usym{2713}$&$\usym{2713}$&$\usym{2717}$&$\usym{2713}$\\
            MGA~\cite{wang2020mgaattack} [2020]&Genetic algorithm&$\usym{2713}$&$\usym{2713}$&$\usym{2713}$&$\usym{2717}$&$\usym{2713}$\\
            MF-GA~\cite{wu2021genetic} [2021]&Genetic algorithm&$\usym{2717}$&$\usym{2713}$&$\usym{2713}$&$\usym{2717}$&$\usym{2713}$\\
            One-Pixel~\cite{su2019one} [2019]&Pixel-wise&$\usym{2717}$&$\usym{2713}$&$\usym{2717}$&$\usym{2713}$&$\usym{2713}$\\
            BruSLe~\cite{vo2024brusleattack} [2024]$^*$&Pixel-wise&$\usym{2717}$&$\usym{2713}$&$\usym{2713}$&$\usym{2713}$&$\usym{2713}$\\ 
            SSAH~\cite{luo2022frequency} [2022]&Frequency domain&$\usym{2713}$&$\usym{2713}$&$\usym{2717}$&$\usym{2713}$&$\usym{2713}$\\
            S$^2$I-FGSM~\cite{long2022frequency} [2022]&Frequency domain&$\usym{2713}$&$\usym{2713}$&$\usym{2717}$&$\usym{2717}$&$\usym{2713}$\\
            Frequency-Aware~\cite{wang2024boosting} [2024]&Frequency domain&$\usym{2713}$&$\usym{2713}$&$\usym{2717}$&$\usym{2717}$&$\usym{2713}$\\
            $N$-HSA$_{LF}$~\cite{cao2024efficient} [2024]&Frequency domain&$\usym{2717}$&$\usym{2713}$&$\usym{2717}$&$\usym{2717}$&$\usym{2713}$\\
            FIA~\cite{wang2021feature} [2021]&Transferable attack&$\usym{2713}$&$\usym{2713}$&$\usym{2717}$&$\usym{2717}$&$\usym{2713}$\\
            DeCoWA~\cite{lin2024boosting} [2024]$^*$&Transferable attack&$\usym{2713}$&$\usym{2713}$&$\usym{2717}$&$\usym{2717}$&$\usym{2713}$\\
            OPS~\cite{guo2025boosting} [2025]$^*$&Transferable attack&$\usym{2713}$&$\usym{2713}$&$\usym{2717}$&$\usym{2717}$&$\usym{2713}$\\
            GAA~\cite{gan2025boosting} [2025]$^*$&Transferable attack&$\usym{2713}$&$\usym{2713}$&$\usym{2717}$&$\usym{2717}$&$\usym{2713}$\\
            LEA2~\cite{qian2023lea2} [2023]&Transferable, Frequency&$\usym{2713}$&$\usym{2713}$&$\usym{2717}$&$\usym{2717}$&$\usym{2713}$\\
            MFI~\cite{qian2024enhancing} [2024]&Transferable, Frequency&$\usym{2713}$&$\usym{2713}$&$\usym{2717}$&$\usym{2717}$&$\usym{2713}$\\
            BASES~\cite{cai2022blackbox} [2022]&Query-and-transfer-based&$\usym{2713}$&$\usym{2713}$&$\usym{2713}$&$\usym{2717}$&$\usym{2713}$\\
            GFCS~\cite{lord2022attacking} [2022]$^*$&Query-and-transfer-based&$\usym{2713}$&$\usym{2713}$&$\usym{2713}$&$\usym{2717}$&$\usym{2713}$\\
            SQBA~\cite{park2024hard} [2024]$^*$&Query-and-transfer-based&$\usym{2713}$&$\usym{2713}$&$\usym{2717}$&$\usym{2717}$&$\usym{2713}$\\
            CG~\cite{feng2022boosting} [2022]&Generator-based&$\usym{2713}$&$\usym{2713}$&$\usym{2713}$&$\usym{2717}$&$\usym{2717}$\\
            MCG~\cite{yin2023generalizable} [2023]$^*$&Generator-based&$\usym{2713}$&$\usym{2713}$&$\usym{2713}$&$\usym{2717}$&$\usym{2717}$\\
            DifAttack~\cite{liu2024difattack} [2024]&Generator-based&$\usym{2713}$&$\usym{2713}$&$\usym{2713}$&$\usym{2717}$&$\usym{2717}$\\
            CDMA~\cite{liu2024boosting} [2024]&Generator-based&$\usym{2713}$&$\usym{2713}$&$\usym{2713}$&$\usym{2717}$&$\usym{2717}$\\
            \midrule
            GreedyPixel (Ours)&\begin{tabular}{c}Query-and-transfer-based\\Pixel-wise\\Greedy algorithm\end{tabular}&$\usym{2713}$&$\usym{2713}$&$\usym{2713}$&$\usym{2713}$&$\usym{2713}$\\
            \bottomrule
        \end{tabular}}
        \begin{tablenotes}
            \item $^*$denotes attacks used for evaluation comparison in \Cref{performance}. ASR denotes attack success rate.
        \end{tablenotes}
    \end{threeparttable}
\end{table*}

\begin{figure}[!t]
    \centering
    \includegraphics[width=\linewidth]{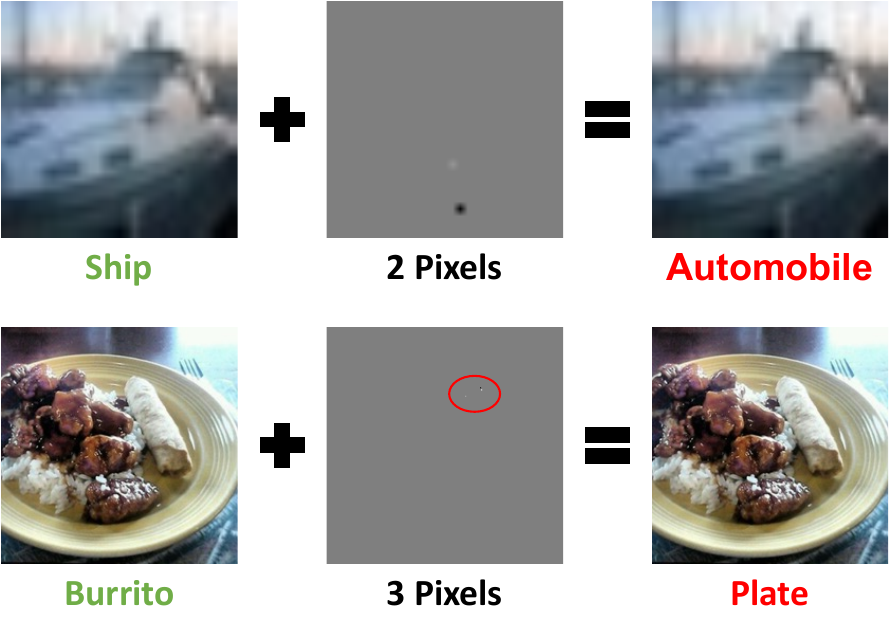}
    \caption{Adversarial examples generated by GreedyPixel, which perturbs only a small, high-saliency subset of pixels. This approach preserves visual fidelity while achieving near white-box attack success rates using only black-box feedback.}
    \label{fig_blackattack}
\end{figure}

\textbf{Gradient access:}
The first dimension distinguishes whether attacks have access to model gradients. \emph{White-box} attacks exploit exact gradients to efficiently solve the optimization problem \cite{goodfellow2014explaining,madry2018towards,croce2020reliable,carlini2017towards,dong2020greedyfool}. These methods are highly effective and query-efficient but rely on access that is rarely available in deployed systems. \emph{Black-box} attacks operate without gradients and instead rely on model outputs, making them more practical but typically less effective and less efficient.

\textbf{Information source:}
The second dimension distinguishes the information source. Within black-box settings, attacks can be divided in accordance with the information they exploit. \emph{Query-only} methods use label or score feedback from the target model \cite{andriushchenko2020square,moon2019parsimonious,wang2020mgaattack,wu2021genetic}. They are flexible and require no additional models but converge slowly and incur a high query cost on high-resolution inputs, and they cannot be directly upgraded if white-box access becomes available. \emph{Transfer-only} methods craft adversarial examples on surrogate models and transfer them to the target \cite{wang2021feature,lin2024boosting,qian2023lea2,qian2024enhancing,guo2025boosting,gan2025boosting}; they are computationally efficient but suffer from limited transferability. \emph{Query-and-transfer} methods combine surrogate guidance with query feedback \cite{cai2022blackbox,lord2022attacking,park2024hard}, mitigating both issues by adapting to the target decision boundary. A key advantage of both transfer-only and query-and-transfer methods is that they can seamlessly leverage white-box information if it becomes available, improving efficiency and attack strength without changing the overall algorithmic framework.

\textbf{Perturbation pattern:}
The third dimension distinguishes how perturbations are distributed across the image. \emph{Pixel-wise or sparse} attacks \cite{su2019one,vo2024brusleattack} modify only a small subset of pixels and can achieve high perceptual quality by concentrating changes on the most influential coordinates. In contrast, \emph{block- or region-based} attacks (\eg Square Attack \cite{andriushchenko2020square}) flip entire patches, which may be more query-efficient, but often produce visually perceptible artifacts and are less precise in localizing the minimal perturbation. \emph{Frequency-domain attacks}~\cite{luo2022frequency,long2022frequency,wang2024boosting,cao2024efficient,qian2023lea2,qian2024enhancing} search for perturbations in the Fourier domain and can improve transferability by exploiting global spectral information, but they typically generate dense, image-wide perturbations that lack fine-grained spatial control and may reduce visual imperceptibility, especially in high-frequency regions where changes are more noticeable.

\textbf{Optimization strategy:}
Finally, the fourth dimension distinguishes whether attacks require additional training. \emph{Direct optimization} methods are training-free and iteratively refine perturbations on-the-fly, enabling immediate adaptation to new targets and threat models. \emph{Training-dependent} approaches include generator-based attacks \cite{yin2023generalizable,liu2024difattack,feng2022boosting,liu2024boosting}, which require large datasets and costly retraining and may overfit to the surrogate data distribution, reducing robustness under domain shift. While generator-based methods produce adversarial examples quickly after training, their training cost and inflexibility limit their applicability in real-time or data-constrained scenarios.

Taken together, these dimensions highlight a gap: no existing attack simultaneously achieves \one query-and-transfer guidance, \two pixel-wise sparsity, and \three training-free direct optimization. To address this gap, we developed GreedyPixel, a fine-grained black-box attack method that combines a surrogate-derived pixel-priority map with greedy, per-pixel optimization refined by query feedback. As illustrated in \Cref{fig_blackattack}, GreedyPixel perturbs only a small set of high-saliency pixels, achieving visually imperceptible yet highly successful attacks without requiring gradients.

As summarized in \Cref{tab_rw}, no prior method jointly satisfies the three criteria mentioned above. Query-only methods lack surrogate guidance and converge slowly; transfer-only methods cannot adapt to the target decision boundary; and generator-based methods require retraining and do not guarantee sparse perturbations. GreedyPixel unifies these three criteria, rendering brute-force-style per-pixel search practical and bridging the precision gap between white-box and black-box attacks.

Our contributions are summarized as follows:
\begin{itemize}
    \item \textbf{Fine-grained brute-force optimization for black-box attacks.} We introduce a surrogate-guided, per-pixel greedy search that makes \emph{brute-force-style} coordinate evaluation practical in black-box settings, delivering truly fine-grained perturbation control without gradient access.
    \item \textbf{White-box-level precision without gradients.} Our deterministic, coordinate-wise updates guarantee monotonic loss decrease and convergence to a local coordinate optimum, enabling precision comparable to white-box attacks while remaining fully training-free and black-box.
    \item \textbf{Comprehensive validation across architectures and threat models.} On the CIFAR-10 and ImageNet datasets, spanning convolutional neural networks (CNNs) and Transformers, GreedyPixel achieved state-of-the-art attack success rates, sparser and less perceptible perturbations, and fast convergence, providing a practical framework for fine-grained robustness evaluation.
\end{itemize}


\section{Related Work}
We review white-box and black-box attacks, summarizing their characteristics and challenges while comparing them with our proposed GreedyPixel method in \Cref{tab_rw}.

\textbf{White-box attacks} rely on full model knowledge and gradient access for perturbation generation. The Fast Gradient Sign Method (FGSM) \cite{goodfellow2014explaining} generates adversarial examples efficiently but often fails to find optimal perturbations. The Projected Gradient Descent (PGD) method \cite{madry2018towards} improves the attack success rate (ASR) through iterative refinement of the perturbation. The Auto PGD (APGD) method \cite{croce2020reliable} adapts step sizes dynamically for better performance, while the Carlini \& Wagner (CW) method \cite{carlini2017towards} achieves higher visual quality at the cost of greater computation. The AutoAttack \cite{croce2020reliable} method integrates multiple algorithms, including APGD. The GreedyFool \cite{dong2020greedyfool} method uses a two-stage greedy strategy but still requires gradient access, whereas the GreedyPixel method uses a gradient-free greedy approach to identify effective perturbations. However, white-box attacks struggle in black-box scenarios due to the unavailability and non-transferability of gradients. Additionally, they often perturb all pixels, producing noticeable differences from the original images.

\textbf{Black-box attacks} generate adversarial examples through diverse strategies, such as single-pixel modification (One-Pixel \cite{su2019one}), submodular search (Square \cite{andriushchenko2020square}, Parsimonious Black-box Attack (PBA) \cite{moon2019parsimonious}), and genetic algorithms (\eg Momentum Gradient Attack (MGA) \cite{wang2020mgaattack}, Multiple Fitness Functions (MF-GA) \cite{wu2021genetic}). The Feature Importance-aware Attack (FIA)~\cite{wang2021feature}, Deformation-Constrained Warping Attack (DeCoWA)~\cite{lin2024boosting}, Orthogonal Projection Strategy (OPS)~\cite{guo2025boosting}, and Ghost Adversarial Attack (GAA)~\cite{gan2025boosting} enhance adversarial example transferability, while methods such as Black-box Attack with Surrogate Ensemble Strategy (BASES) \cite{cai2022blackbox}, Greedy-Fool Coordinate Search (GFCS) \cite{lord2022attacking}, and Small Query Black-box Attack (SQBA) \cite{park2024hard} integrate surrogate gradients with query feedback and thereby achieve higher success rates. The Query-Efficient Score-Based Black-Box Sparse Adversarial Attack (BruSLe) \cite{vo2024brusleattack} discovers sparse perturbations through Bayesian optimization.

Frequency-domain approaches (S$^2$I-FGSM \cite{long2022frequency}, $N$-HSA$_{LF}$ (HSA: harmonic spectrum alignment) \cite{cao2024efficient}, Frequency-Aware \cite{wang2024boosting}, SSAH (Semantic Similarity Attack on High-frequency components) \cite{luo2022frequency}) modify spectral components to improve transferability. Recent methods such as LEA2 (Lightweight Ensemble Adversarial Attack (via Non-overlapping Vulnerable Frequency Regions)) \cite{qian2023lea2} and MFI (Mixed-Frequency Input) \cite{qian2024enhancing} further exploit frequency subspaces by targeting non-overlapping vulnerable bands or by mixing frequency components across images, respectively, and achieve strong transfer results on both CNNs and Transformers. However, frequency-domain methods typically generate dense perturbations that are difficult to constrain for sparsity, and they cannot adapt to the target decision boundary without additional queries.

Generator-based methods (Conditional Generative Attack (CG) \cite{feng2022boosting}, Meta-learning-based Conditional Generator (MCG) \cite{yin2023generalizable}, DifAttack \cite{liu2024difattack}, Conditional Diffusion Model Attack (CDMA) \cite{liu2024boosting}) use learned feature spaces to generate adversarial examples with a single forward pass but require expensive retraining and may overfit to the training distribution.

Overall, existing black-box attacks face a high query cost, limited adaptability to white-box information, a lack of per-instance optimality, and weak control of perturbation sparsity; many rely on random search, gradient approximation, or region-/frequency-based sampling, which restricts precision. In contrast, \emph{GreedyPixel} performs \emph{brute-force-style, per-pixel greedy optimization} that directly evaluates each coordinate without gradients, thereby achieving precise fine-grained control. Coupled with a surrogate-driven pixel-priority map and query-refined updates, this design guarantees monotonic loss reduction, coordinate-wise convergence, and explicit sparsity control, thereby approaching the precision of gradient-based white-box methods while remaining fully black-box and training-free.

\section{GreedyPixel Methodology}
\label{method}
This section presents the GreedyPixel attack methodology. \Cref{threatmodel,problem} define the threat models and problem formulation. \Cref{overview} introduces the key design principles and underlying rationale, situating GreedyPixel among related attack families. The attack constructs a pixel-wise priority map (\Cref{map}) and iteratively processes pixels (\Cref{greedy}). Finally, the complete algorithm is summarized in \Cref{algorithm}.

\subsection{Threat Model}
\label{threatmodel}
GreedyPixel is primarily designed for black-box attacks but adapts to white-box scenarios when gradient access is available. In black-box mode, it relies on query feedback, using a surrogate model to prioritize pixels. In white-box mode, it directly utilizes target model gradients, streamlining perturbation generation. Unlike most black-box attacks, GreedyPixel can transition to a white-box attack, as transfer-based methods allow this adaptation. Additionally, it operates without $L_\infty$-norm constraints, similar to sparse attacks like One-Pixel \cite{su2019one} and BruSLe \cite{vo2024brusleattack}. GreedyPixel supports three threat models:
\begin{itemize}
    \item White-box with constrained $L_\infty$-norm;
    \item Black-box with constrained $L_\infty$-norm;
    \item Black-box without $L_\infty$-norm constraints.
\end{itemize}

\subsection{Adversary Goal and Loss Function}
\label{problem}
Adversarial attacks aim to manipulate predictions through subtle perturbations. Let $x$ and $y$ denote the original image and its ground-truth label, respectively, and let $x'$ represent the adversarial example, formulated as
\begin{equation}
    x'=x+\delta,\quad \text{s.t.}\ \|\delta\|_\infty\leq\epsilon,
\end{equation}
where $\epsilon$ is the maximum allowable perturbation magnitude. Thus, given a loss function $\ell$ with respect to $x'$, the optimization objective in GreedyPixel to achieve misclassification is
\begin{equation}
    x'=\arg\min_{x'}\ell(x').
    \label{eq_optimization}
\end{equation}

GreedyPixel utilizes the CW loss function \cite{carlini2017towards} to define its optimization problem. The CW loss is designed to alter the prediction between the target and highest possible non-target label. For the misclassification scenario, the loss function used by GreedyPixel is formulated as
\begin{equation}
    \begin{aligned}
        \ell(x')=Z(x')_y-\max_{i\neq y}Z(x')_i,
    \end{aligned}
    \label{eq_loss}
\end{equation}
where $Z(x')$ represents the logits (pre-softmax activations) of the target model (for brevity, we use $Z$ to refer to the target model), and $y$ and $i$ denote the indices, referring to labels. Specifically, $\ell \geq 0$ indicates a correct classification, while $\ell<0$ indicates a successful misclassification attack.

\subsection{Design Overview and Positioning}
\label{overview}
To clearly position GreedyPixel and highlight its novelty, we clarify its design principle and contrast it with related attacks. Our design is fundamentally motivated by brute-force search, which is optimal but computationally infeasible. The following discussion formalizes how we systematically reduce the complexity of brute-force search to a practical level and highlights the resulting algorithmic advantages.

\paragraph{Brute-force motivation}
Brute-force is optimal but infeasible for practical image sizes. For an image of size $C \times H \times W$, the continuous perturbation space is unbounded:
\begin{equation}
    Q_{\text{continuous}} = \mathcal{O}(\infty^{H \times W \times C}),
\end{equation}
where each pixel may take infinitely many values. Restricting perturbations to the discrete range $\{-\epsilon,\ldots,+\epsilon\}$ yields
\begin{equation}
  \begin{aligned}
     Q_{\text{discrete}} &= \{-\epsilon,\ldots,+\epsilon\}^{H \times W \times C}\\
     &= \bigl(2 \times \text{int}(\epsilon \cdot 255) + 1\bigr)^{H \times W \times C},
  \end{aligned}
\end{equation}
where $\text{int}(\epsilon \cdot 255)$ denotes the maximum pixel change on a 1--255 scale, ``2'' accounts for $\pm$, and ``$+1$'' includes 0. Even if we restrict the perturbations to only the most extreme values ($-\epsilon$ and $+\epsilon$), the space remains exponential:
\begin{equation}
    Q_{\text{binary}} = \{-\epsilon,+\epsilon\}^{H \times W \times C} = 2^{H \times W \times C}.
    \label{eq_bruteforce_binary}
\end{equation}

\textbf{This binary restriction is justified.} In our analysis of 1,000 adversarial examples generated by AutoAttack \cite{croce2020reliable}, 70.43\% of the  perturbations in $L_\infty$-norm attacks occurred at $\pm \epsilon$, confirming the importance of these extreme values. Therefore, GreedyPixel is designed as a principled reduction of brute-force search, focusing on $\pm \epsilon$ perturbations and systematically reducing complexity to a tractable pixel-wise form.

\paragraph{Fine-grained vs. coarse-grained search}
We define \emph{fine-grained} attacks as those that operate directly at the pixel level (\eg PGD~\cite{madry2018towards}), while \emph{coarse-grained} attacks update regions or blocks (\eg Square Attack~\cite{andriushchenko2020square}). Coarse-grained updates often lead to suboptimal perturbations because their search space $\mathcal{S}_{\text{coarse}}$ is a strict subset of the full pixel-wise space $\mathcal{S}_{\text{pixel}}$:
\begin{equation}
    \mathcal{S}_{\text{coarse}} \subset \mathcal{S}_{\text{pixel}}.
\end{equation}
Thus, even with infinite queries, coarse-grained search cannot guarantee global optimality. GreedyPixel, in contrast, deterministically exhausts the per-pixel candidate space in a fixed ordering. GreedyPixel is the first query-based black-box attack to implement a fine-grained brute-force strategy without gradient access.

\paragraph{Local exhaustive search vs. gradient-based local optima}
Instead of enumerating all pixels jointly, GreedyPixel applies brute-force locally to each pixel in sequence. The per-pixel cost is
\begin{equation}
    Q_{\text{pixel}} = \{-\epsilon,+\epsilon\}^{C} = 2^{C},
\end{equation}
which is trivial. Running over all pixels once (a one-pass search) requires
\begin{equation}
    Q_{\text{one-pass}} = H \times W \times 2^{C},
    \label{eq_onepass}
\end{equation}
which transforms an exponential search into a linear-time procedure with respect to the number of pixels. This makes repeated passes computationally feasible even under query budgets (\eg $Q_{\text{one-pass}} = 8192$ on CIFAR-10). Repeated passes approximate coverage of the full binary space $Q_{\text{binary}}$ (\Cref{eq_bruteforce_binary}) in a sequential manner. Unlike gradient-based attacks (\eg CW~\cite{carlini2017towards}, PGD~\cite{madry2018towards}), which may become trapped in local minima due to following a fixed gradient direction \cite{bengio1994learning,hinton2006fast}, GreedyPixel guarantees local optimality at each per-pixel update and empirically converges towards solutions close to the global optimum with repeated passes.

\paragraph{Surrogate-guided ordering}
Although pixel-wise exhaustive search is feasible, iterating over all pixels uniformly is inefficient under query constraints. To accelerate convergence, GreedyPixel leverages a surrogate model to prioritize pixels in accordance with their importance. We specifically use adversarially trained surrogates that emphasize robust, semantically meaningful features such as object edges and textures rather than spurious background correlations \cite{tramer2018ensemble}. Perturbations placed on these high-saliency regions are less perceptually noticeable than those on flat backgrounds, aligning with evidence from human perception that distortions are masked in textured regions. Our ablation confirms that adversarially trained surrogates outperform non-robust surrogates and random orderings in both efficiency and perceptual quality.

\paragraph{Training-free optimization}
GreedyPixel is training-free and iteratively refines perturbations in input space, guaranteeing immediate applicability to any new model or threat setting. This strategy is highly flexible and naturally supports both targeted and untargeted objectives. By operating directly on model outputs, it avoids overfitting and remains robust even when the target model distribution differs from that of any surrogate or auxiliary dataset.

In contrast, training-dependent approaches include generator-based attacks and learned surrogates \cite{yin2023generalizable,liu2024difattack,feng2022boosting,liu2024boosting}. While they can generate adversarial examples with a single forward pass at inference time, they require substantial labeled data and computational resources for training, and must also be retrained whenever the target model or task changes. This additional retraining overhead makes them impractical in dynamic or online settings where the model evolves or defenses are updated.

\begin{figure*}[!t]
    \centering
    \includegraphics[width=6in]{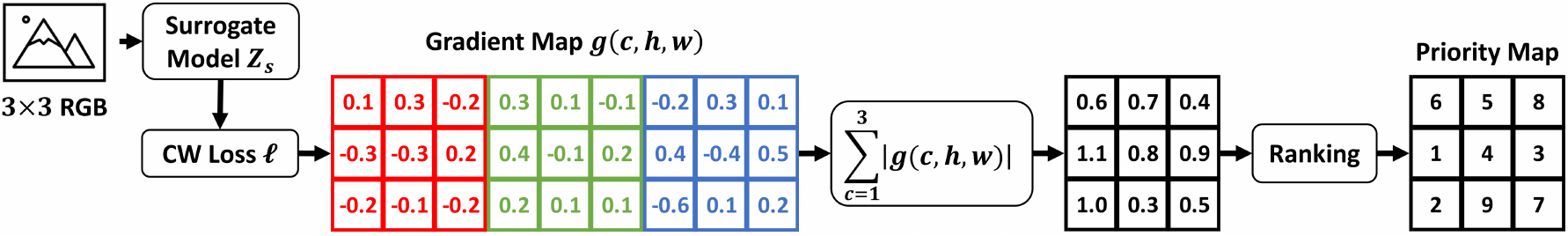}
    \caption{Pixel-wise priority map obtained by ranking gradient $g$ of the CW loss function $\ell$ with respect to input $x$ and ground-truth label $y$ within a surrogate model $Z_s$. Replacing the surrogate model with the target model yields a white-box attack.}
    \label{fig_prioritymap}
\end{figure*}

Moreover, generators optimize an expected loss over the training distribution and cannot guarantee per-instance optimality, often resulting in suboptimal perturbations or reduced transferability under distribution shift. In addition, their dense, image-wide perturbations make it challenging to enforce sparsity constraints or guarantee visual imperceptibility, whereas direct optimization can explicitly control sparsity at the level of individual pixels.

These limitations motivate our adoption of direct, training-free optimization, an approach that is inherently adaptable, data-efficient, and theoretically grounded in a coordinate-wise greedy search strategy.

\paragraph{Comparison to prior work}
\begin{itemize}
    \item \textbf{Gradient-based white-box attacks} (\eg CW~\cite{carlini2017towards}, PGD~\cite{madry2018towards}) follow gradients and thus risk getting trapped in local minima; GreedyPixel avoids this through exhaustive per-pixel updates.
    \item \textbf{Greedy algorithms} (\eg GreedyFool~\cite{dong2020greedyfool}) greedily select pixel subsets for gradient descent and thus remain gradient-dependent. GreedyPixel dispenses with gradients entirely.
    \item \textbf{Pixel-wise attacks} (\eg One-Pixel~\cite{su2019one}, BruSLe~\cite{vo2024brusleattack}) operate at the pixel level but search a restricted perturbation space. GreedyPixel evaluates all $2^C$ channel perturbations, covering a superset of states.
    \item \textbf{Transfer-based attacks} (\eg DeCoWA~\cite{lin2024boosting}) are query-free but lack robustness when faced with surrogate–target mismatch. GreedyPixel actively adapts through queries and thereby achieves stronger success rates.
    \item \textbf{Random-search attacks} (\eg Square Attack~\cite{andriushchenko2020square}) perturb coarse blocks stochastically and explore only a restricted subset of the pixel-wise space.
    \item \textbf{Generator-based attacks} (\eg MCG~\cite{yin2023generalizable}) enable fast inference after training but require substantial labeled data and retraining for each new target, do not guarantee per-instance optimality, and struggle to enforce sparsity or maintain visual imperceptibility under distribution shift.
\end{itemize}
Together, these differences establish GreedyPixel as the first practical, query-based, fine-grained brute-force attack method that is both robust and perceptually aware.

\begin{tcolorbox}[takeaway,title=Takeaway]
GreedyPixel performs local exhaustive pixel-wise search guided by surrogate-based prioritization and refined through query feedback. This approach \one reduces brute-force search to linear time, \two avoids gradient-based local minima, \three concentrates perturbations on robust, semantically meaningful pixels to preserve visual quality, and \four is training-free, eliminating generator retraining costs while remaining directly upgradeable to white-box settings.
\end{tcolorbox}

\subsection{Pixel-Wise Priority Map}
\label{map}
The pixel-wise priority map determines the order in which pixels are updated by GreedyPixel and is computed entirely on the basis of the surrogate model $Z_s$ (used interchangeably for model and logits). Each pixel's priority is derived from the gradient of the CW loss $\ell$ with respect to input $x$, so pixels with the greatest potential saliency are perturbed first. Because this process relies only on $Z_s$, it requires no queries to the target model. When $Z_s$ equals the target model (white-box), the same ordering yields the fastest convergence, and early stopping minimizes the number of manipulated pixels.

As shown in \Cref{fig_prioritymap}, the priority map is generated as:
\begin{itemize}
    \item \textit{Step 1:} Compute $g(c,h,w)=\nabla_x \ell(x)$ on $Z_s$, where $c,h,w$ denote channel, height, and width.
    \item \textit{Step 2:} Aggregate per-pixel importance $g'(h,w)=\sum_{c=1}^C |g(c,h,w)|$.
    \item \textit{Step 3:} Sort pixels by $g'(h,w)$ in descending order to form the update sequence.
\end{itemize}
\Cref{alg_map} summarizes the procedure.

\begin{algorithm}[!t]
    \caption{Pixel-Wise Priority Map}
    \label{alg_map}
    \begin{algorithmic}[1]
        \Require{Original image $x$ of size $C \times H \times W$ with ground-truth label $y$ and surrogate model $Z_s$.}
        \vspace{0.1cm}
        \Ensure{Priority map $\{(h_i,w_i)\}_{i=1}^{H \times W}$.}
        \vspace{0.1cm}
        \State \(\ell(x)=Z_s(x)_y-\max_{i\neq y}Z_s(x)_i\)
        \vspace{0.1cm}
        \State \(g(c,h,w)=\nabla_x\ell(x)\)\Comment{gradient map}
        \vspace{0.1cm}
        \State \(g'(h,w)=\sum_{c=1}^C|g(c,h,w)|\)\Comment{pixel importance}
        \vspace{0.2cm}
        \State \(\bigl\{(h_i, w_i)\bigr\}_{i=1}^{H \times W} = \operatorname{Sort}\bigl(g'(h,w)\bigr)_{\text{desc}}\)
        \vspace{0.1cm}
        \State \Return{\(\{(h_i,w_i)\}_{i=1}^{H \times W}\)}
    \end{algorithmic}
\end{algorithm}

\textbf{Theoretical Rationale.}
We formalize the advantages of using a priority map by means of three complementary arguments:

\paragraph{Coverage without redundancy}
Let $H$ and $W$ denote the image height and width, respectively, so the total number of pixel positions is $M = H \times W$. If pixel positions are sampled uniformly \emph{with replacement}, the expected number of selections required to visit all $M$ positions at least once (one-pass) is given by the coupon collector result~\cite{motwani1995randomized}:
\begin{equation}
    \mathbb{E}[\text{one-pass}] = M \cdot H_{M} \approx M \ln M,
    \label{eq_coupon}
\end{equation}
where $H_{M}$ is the $M$-th harmonic number $H_{M} = \sum_{k=1}^{M} \frac{1}{k}$. This shows that many queries are wasted revisiting pixels before one-pass coverage is achieved. By contrast, any fixed priority map (even a random but fixed permutation of the $M$ pixels) guarantees coverage in exactly $M$ steps, eliminating redundancy and ensuring that each pixel is processed exactly once per pass.

\paragraph{White-box optimality} Let $g(p) = \nabla_{x_p}\ell(x)$ denote the gradient of the loss $\ell$ with respect to pixel $p$ for RGB values ($p \in \{1,\dots,M\}$), and let $g_s(p)$ be the corresponding gradient on the surrogate model $Z_s$. When the surrogate matches the target (\ie $Z_s = Z$), we have $g_s(p) = g(p)$ for all pixels, and ranking pixels by $\|g_s(p)\|_1$ corresponds to performing steepest coordinate descent on the target loss. The locally optimal update for a single pixel (by first-order Taylor expansion) yields approximately
\begin{equation}
    \Delta\ell_p \approx -\epsilon \|g(p)\|_1 = -\epsilon \|g_s(p)\|_1.
    \label{eq_white}
\end{equation}
Processing pixels in descending order of $\|g_s(p)\|_1$ therefore achieves the maximum per-step loss reduction and yields the fastest convergence. Early stopping once misclassification is achieved further minimizes the number of perturbed pixels.

\paragraph{Surrogate quality and perceptual benefit}
When $Z_s \neq Z$, ranking pixels by $\|g_s(p)\|_1$ remains superior to random selection as long as the surrogate and target pixel-wise gradients are positively aligned. We define their cosine similarity at pixel $p$ as
\begin{equation}
    \cos\theta_p = \frac{g_s(p)^\top g(p)}{\|g_s(p)\|_2 \|g(p)\|_2}.
    \label{eq_model_cosine}
\end{equation}
If $\cos\theta_p > 0$, the target-loss reduction per-pixel update tends to increase with
\begin{equation}
    -\Delta\ell_p \propto \epsilon \cdot \rho_p \cdot \|g_s(p)\|_1,
    \label{eq_black}
\end{equation}
where $\rho_p = \cos\theta_p$ quantifies the alignment at pixel $p$. Thus, pixels with larger $\|g_s(p)\|_1$ are expected to yield greater loss reduction on the target, even when $Z_s \neq Z$.

Adversarially trained surrogates increase $\rho_p$ because they produce gradients that emphasize robust, semantically meaningful features such as object edges and textures~\cite{tramer2018ensemble}. Prioritizing these regions benefits both attack performance and perceptual quality: it improves ASR by perturbing genuinely predictive features while reducing perceptual distortion because perturbations are placed on visually complex regions rather than smooth backgrounds, thereby aligning with texture-masking effects in human vision.

\begin{tcolorbox}[takeaway,title=Takeaway]
The pixel-wise priority map is theoretically grounded in three complementary properties: \one it guarantees efficient one-pass coverage without redundant queries; \two it achieves optimal convergence speed when the surrogate matches the target; \three it remains superior to random ordering even when the surrogate and target do not match by prioritizing robust, semantically meaningful pixels. Together, these properties result in faster attacks and visually subtler adversarial examples.
\end{tcolorbox}

\subsection{Pixel-Wise Greedy Algorithm}
\label{greedy}
This subsection introduces the core mechanism of GreedyPixel. We first present the per-pixel greedy update rule, then show how score-based query feedback ensures a monotonic decrease in the target loss, and finally analyze convergence and its connection to global optimality.

\subsubsection{Greedy Update Rule}
Let $\delta^{(t)} \in \mathbb{R}^{C \times H \times W}$ denote the full perturbation tensor at iteration $t$, so that
\begin{equation}
    x^{(t)} = x + \delta^{(t)}.
\end{equation}
At each step, a pixel $p = (h,w)$ is selected from the priority map produced by \Cref{alg_map}. Let $\delta^{(t)}[p] \in \mathbb{R}^C$ denote the current $C$-dimensional perturbation vector at $p$, and define the zeroed perturbation as
\begin{equation}
    \delta^{(t)}_{-p} = \delta^{(t)} - \mathcal{I}_p\bigl(\delta^{(t)}[p]\bigr),
\end{equation}
where $\mathcal{I}_p(\cdot)$ embeds a $C$-channel vector at pixel $p$ (zeros elsewhere). The algorithm exhaustively enumerates all $2^C$ candidate vectors $\delta_p \in \{-\epsilon,+\epsilon\}^C$ and selects
\begin{equation}
    \hat{\delta}_p = \arg\min_{\delta_p \in \{-\epsilon,+\epsilon\}^C} \ell\bigl(x + \delta^{(t)}_{-p} + \mathcal{I}_p(\delta_p)\bigr).
\end{equation}
The pixel update is applied in accordance with
\begin{equation}
    \delta^{(t+1)} = \delta^{(t)}_{-p} + \mathcal{I}_p(\hat{\delta}_p), \qquad
    x^{(t+1)} = x + \delta^{(t+1)}.
\end{equation}
Thus, each iteration perturbs exactly one pixel with the locally optimal choice, leaving all other pixels fixed. The complete procedure for the greedy strategy at a single pixel is provided in \Cref{alg_greedy}.

\begin{algorithm}[!t]
    \caption{Greedy Strategy for a Single Pixel (with Query Feedback)}
    \label{alg_greedy}
    \begin{algorithmic}[1]
        \Require{Image $x$ of size $C \times H \times W$, current perturbation $\delta^{(t)}$, selected pixel $p$, target model $Z$, loss function $\ell(\cdot)$, perturbation set $\{-\epsilon,+\epsilon\}$.}
        \Ensure{Updated perturbation $\delta^{(t+1)}$, updated loss $\ell^{(t+1)}$.}
        \State $\delta^{(t)}_{-p} = \delta^{(t)} - \mathcal{I}_p(\delta^{(t)}[p])$ \Comment{zero out current pixel}
        \For{each $\delta_p \in \{-\epsilon,+\epsilon\}^C$} 
            \State $\tilde{\delta} = \delta^{(t)}_{-p} + \mathcal{I}_p(\delta_p)$
            \State $\ell_{\delta_p} = \ell(x + \tilde{\delta})$
        \EndFor
        \State $\hat{\delta}_p = \arg\min_{\delta_p} \ell_{\delta_p}$ \Comment{select optimal candidate}
        \If{$\ell_{\hat{\delta}_p} < \ell(x+\delta^{(t)})$} \Comment{accept if loss decreases}
            \State $\delta^{(t+1)} = \delta^{(t)}_{-p} + \mathcal{I}_p(\hat{\delta}_p)$
        \Else
            \State $\delta^{(t+1)} = \delta^{(t)}$ \Comment{reject, keep unchanged}
        \EndIf
        \State $\ell^{(t+1)} = \ell(x+\delta^{(t+1)})$
        \State \Return{$\delta^{(t+1)}, \ell^{(t+1)}$}
    \end{algorithmic}
\end{algorithm}

\subsubsection{Score-Based Query Feedback}
Surrogate--target alignment may be imperfect (\eg robust surrogate \vs non-robust target), and relying purely on surrogate ordering can yield poor transferability, as evidenced by DeCoWA results~\cite{lin2024boosting} (\Cref{tab_cifar10,tab_rn50,tab_vit,tab_vgg}). GreedyPixel mitigates this by incorporating \emph{score-based query feedback}: the candidate update $\hat{\delta}_p$ is committed only if it strictly reduces the target loss,
\begin{equation}
    \ell\bigl(x + \delta^{(t)}_{-p} + \mathcal{I}_p(\hat{\delta}_p)\bigr)
    < \ell\bigl(x^{(t)}\bigr).
\end{equation}
Otherwise, the pixel is left unchanged ($\delta^{(t+1)} = \delta^{(t)}$, Step 10 of \Cref{alg_greedy}). This constitutes an exact coordinate-descent step with a discrete line search along coordinate $p$. Because each accepted update reduces the loss, $\{\ell(x^{(t)})\}$ is monotonically non-increasing and therefore convergent.

\textbf{Why surrogate ordering + feedback outperforms transfer-only?}
A purely transfer-based attack relies on the assumption of positive alignment,
\begin{equation}
    \cos\theta_p = 
    \frac{g_s(p)^\top g(p)}{\|g_s(p)\|_2 \|g(p)\|_2} > 0,
\end{equation}
so that $\mathrm{sign}\bigl(g_s(p)\bigr)$ serves as a reliable proxy for $\mathrm{sign}\bigl(g(p)\bigr)$. When $\cos\theta_p$ is near zero or negative, the surrogate may produce harmful updates that increase the loss \cite{tramer2018ensemble}. Target model feedback corrects instances of misalignment by rejecting such updates, while preserving effective updates under strong surrogate–target alignment (Steps 7--11 of \Cref{alg_greedy}). Thus, combining surrogate ordering with target feedback yields efficiency when alignment is strong and robustness when the surrogate and target are mismatched.

\begin{tcolorbox}[takeaway,title=Takeaway]
Score-based feedback enables GreedyPixel to act as an adaptive algorithm that guarantees a monotonic decrease in the target loss. Surrogate ordering enables it to achieve fast convergence when alignment is strong and robust performance when alignment is weak, resulting in a higher ASR and lower perceptual distortion within the same query budget.
\end{tcolorbox}

\subsubsection{Convergence and Global Optimality Approximation}
After one full pass over all $M = H \times W$ pixels, no single-pixel flip can further reduce the loss:
\begin{equation}
    \ell(x^{(T+1)}) \le 
    \ell(x^{(T)} + \mathcal{I}_p(\delta_p)), \quad
    \forall p,\ \delta_p \in \{-\epsilon,+\epsilon\}^C.
\end{equation}
Thus, $x^{(T+1)}$ is a coordinate-wise local minimum in the discrete space $\{-\epsilon,+\epsilon\}^{C\times M}$. Repeated passes progressively refine the solution, and, as $T \to \infty$, the algorithm approaches a deeper coordinate-wise minimum; empirically, this is often close to the global optimum of the discrete space.

\begin{figure}[!t]
    \centering
    \includegraphics[width=3.0in]{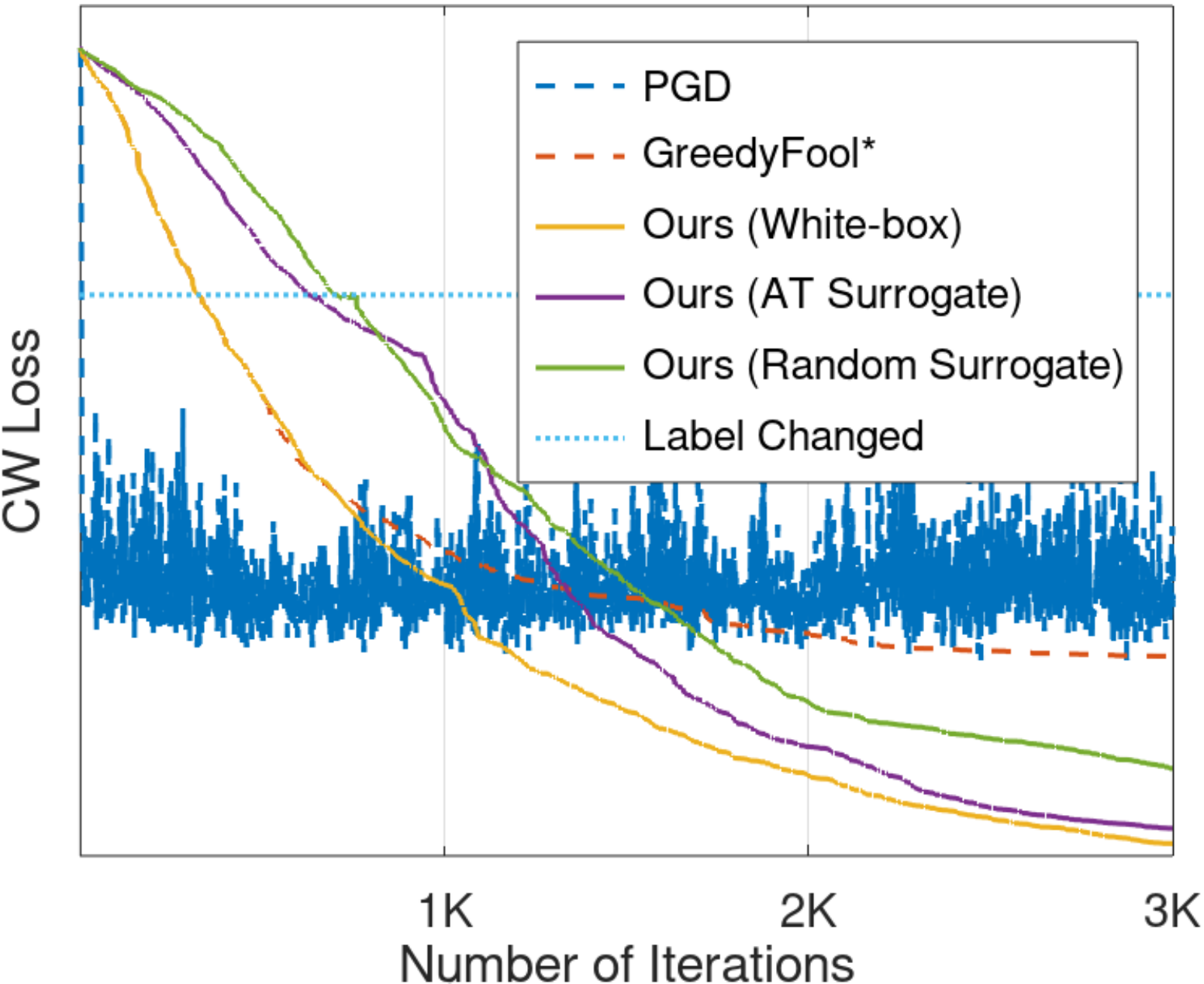}
    \caption{CW loss \vs number of iterations for PGD~\cite{madry2018towards}, GreedyFool$^*$, and GreedyPixel variants. GreedyFool$^*$ uses GreedyFool's pixel-selection heuristic (top-50\% salient pixels) \cite{dong2020greedyfool} but replaces its gradient-descent update with a pixel-wise exhaustive search, ensuring a fair comparison under black-box assumptions. However, GreedyFool$^*$ converges to a higher loss plateau, revealing the limitations of its pixel-selection strategy. GreedyPixel achieves the lowest final loss, with white-box access converging fastest, followed by adversarially trained (AT) surrogate and random surrogate settings. The dotted line indicates the first-iteration label change.}
    \label{fig_convergence}
\end{figure}

\textbf{Avoidance of gradient-based local minima.}
Gradient-based attacks such as PGD~\cite{madry2018towards} update all pixels jointly:
\begin{equation}
    x^{(t+1)} = x^{(t)} - \alpha \cdot \mathrm{sign}\bigl(\nabla_x \ell(x^{(t)})\bigr).
\end{equation}
On rugged loss landscapes, $\nabla_x \ell$ may vanish or guide the optimization into suboptimal local basins, causing premature convergence \cite{bengio1994learning,hinton2006fast}. GreedyPixel circumvents this pitfall by exhaustively evaluating candidate flips for each pixel, ensuring every accepted update strictly reduces the loss. As illustrated in \Cref{fig_convergence}, GreedyPixel achieves lower loss than PGD and GreedyFool$^*$ even under an identical pixel-selection budget.

\begin{tcolorbox}[takeaway,title=Takeaway]
GreedyPixel converges monotonically, achieves coordinate-wise optimality after each pass, and asymptotically approaches the global optimum. Unlike PGD, it is not susceptible to traps from local gradient directions and thus remains effective even on rugged, non-smooth loss landscapes.
\end{tcolorbox}

\subsection{Full Pipeline of GreedyPixel Attack}
\label{algorithm}
Throughout its iterations, GreedyPixel (\Cref{alg_GreedyPixel}) processes pixels in accordance with the priority map (\Cref{alg_map}) and applies the greedy update rule with query feedback (\Cref{alg_greedy}). Additionally, the pixel-priority order may be refreshed periodically every $N$ iterations to concentrate queries on the most promising coordinates while keeping the perturbation sparse. The attack terminates when either \one the maximum query budget is reached, $t \cdot 2^C \leq Q_\text{max}$, or \two misclassification is achieved, $\ell(x^{(t)}) < 0$.

The total query cost is thus
\begin{equation}
    Q_{\mathrm{Greedy}} = t \times 2^C \le Q_\text{max},
    \label{eq_querycost}
\end{equation}
where $2 = |\{-\epsilon,+\epsilon\}|$. This illustrates the efficiency of GreedyPixel relative to brute-force search ($2^{C\times H \times W}$ evaluations, \Cref{eq_bruteforce_binary}). When $t \geq H \times W$, every pixel has been processed at least once. Thus, the number of manipulated pixels is $\min(t, H \times W)$, which scales with image resolution and directly corresponds to the query cost.

\begin{algorithm}[!t]
    \caption{GreedyPixel Attack (Full Pipeline)}
    \label{alg_GreedyPixel}
    \begin{algorithmic}[1]
        \Require{Input $x$ of size $C \times H \times W$, label $y$, target model $Z$, surrogate model $Z_s$, maximum perturbation $\epsilon$, maximum query budget $Q_\text{max}$, priority-map refresh period $N$.}
        \State $\delta^{(0)} = \mathbf{0}$, $t=0$
        \While{$t \cdot 2^C \le Q_\text{max}$}
            \If{$t \bmod N = 0$}
                \State $\{p_i\}_{i=1}^{H \times W} = \Cref{alg_map}(x,y,Z_s)$\Comment{refresh priority map}
            \EndIf
            \State $p = p_{(t \bmod H \times W)}$
            \State $\delta^{(t+1)},\ell^{(t+1)} = \Cref{alg_greedy}(x,y,Z,p,\delta^{(t)},\epsilon)$
            \If{$\ell^{(t+1)} < 0$}
                \State \textbf{break}
            \EndIf
            \State $t \gets t+1$
        \EndWhile
        \State \Return $x' = x + \delta^{(t+1)}$
    \end{algorithmic}
\end{algorithm}

\section{Experiment Settings}
\subsection{Datasets, Target, and Surrogate Models}
Attacks were evaluated on the CIFAR-10 \cite{krizhevsky2009learning} (32$\times$32) and ImageNet \cite{nips2017} datasets (cropped to $64 \times 64$ and $224 \times 224$ for specific studies), each with 1,000 examples. The target models included four non-robust pretrained models: WideResNet-28-10 \cite{croce2021robustbench} for CIFAR-10, and ResNet-50 \cite{croce2021robustbench}, ViT-Base-Patch16 \cite{wu2020visual}, and VGG19-BN \cite{simonyan2015very} for ImageNet. The inclusion of the Vision Transformer model (ViT-Base-Patch16) enabled a direct assessment of GreedyPixel on a non-CNN architecture and illustrates its generalizability beyond convolutional networks. Three robust models were also tested: WideResNet-28-10 \cite{wang2023better}, WideResNet-82-8 \cite{bartoldson2024adversarial}, and XCiT-S12 \cite{debenedetti2023light}.

Ablation studies were conducted using various surrogate models \cite{wong2020fast,gowal2021improving,peng2023robust,singh2023revisiting,liu2023comprehensive,salman2020adversarially}, with fairness ensured by using the same surrogate model across attacks. In black-box settings, the surrogate differed from the target, while in white-box settings, they were identical. GreedyPixel utilized adversarially pretrained surrogate models for improved visual quality, whereas non-adversarial surrogate models were approximated using random sequences.

\subsection{Benchmarks and Experimental Settings}
\label{settings}
The proposed GreedyPixel method was benchmarked against leading adversarial attack techniques under the three threat models described in \Cref{threatmodel}, with specific attacks selected from various strategies summarized in \Cref{tab_rw}.

\textbf{White-box attacks with constrained $L_\infty$-norm distance (\ie~limited $\epsilon$):} GreedyPixel was compared with state-of-the-art gradient-based white-box attacks, including AutoAttack \cite{croce2020reliable} and APGD \cite{croce2020reliable}, under a uniform perturbation constraint of $\epsilon = 4/255$. In this evaluation, the number of queries was left unrestricted.

\textbf{Black-box attacks with constrained $L_\infty$-norm distance (\ie~limited $\epsilon$):} GreedyPixel was evaluated against several prominent black-box attack methods, including DeCoWA \cite{lin2024boosting} (purely transfer-based), GFCS \cite{lord2022attacking} (score-and-transfer-based), MCG \cite{yin2023generalizable} (generator-based), SQBA \cite{park2024hard} (score-and-transfer-based), and two recent attack methods (OPS~\cite{guo2025boosting} and GAA~\cite{gan2025boosting}), all under the same $\epsilon = 4/255$ constraint. The methods were evaluated under a maximum query budget of 20,000.

\textbf{Black-box attacks with unconstrained $L_\infty$-norm distance (\ie~unlimited $\epsilon$):} For this setting, we compared GreedyPixel with a state-of-the-art sparse attack method, BruSLe \cite{vo2024brusleattack}, which has shown superior ASRs compared with previous methods like One-Pixel attack \cite{su2019one}. BruSLe’s sparsity was fixed at 1\% (10 pixels) for CIFAR-10 and 0.1\% (50 pixels) for ImageNet, with both methods operating under a maximum query budget of 20,000.

\textbf{Against adversarial defenses:} The attacks were further evaluated against established adversarial defense mechanisms, including a state-of-the-art adversarial purification method, DiffPure \cite{nie2022diffusion}, three recent adversarial training (AT) defenses \cite{wang2023better, bartoldson2024adversarial, debenedetti2023light}, and one adversarial detection method \cite{meng2017magnet}, across all three threat models.

\subsection{Evaluation Metrics}
The evaluation metrics included several key aspects to assess the performance of the proposed attack and benchmark methods. An attack's effectiveness was measured by its ASR, with higher values indicating greater success in manipulating model predictions. Visual quality was assessed using the Structural Similarity Index Measure (SSIM) \cite{wang2004image} and Learned Perceptual Image Patch Similarity (LPIPS) \cite{zhang2018unreasonable}, where higher SSIM and lower LPIPS values indicate subtler deviations from the original images. Computational efficiency was measured by execution time, with shorter durations indicating more efficient attacks. The average number of queries (Avg. Q) reflected query efficiency, with fewer queries indicating better resource utilization. The adversarial detection rate quantified the proportion of adversarial examples identified as threats and was included as part of the assessment of attacks against adversarial defenses. A lower detection rate indicates greater attack effectiveness, reflecting better evasion of detection.

\subsection{Implementation}
GreedyPixel was implemented using the Adversarial Robustness Toolbox \cite{art2018}, with RobustBench \cite{croce2021robustbench} serving as the benchmark for evaluating adversarial robustness. The experiments were conducted on a workstation equipped with an NVIDIA A100 40-GB GPU.

\section{Performance}
\label{performance}

\subsection{Attack Effectiveness}
We evaluated the GreedyPixel attack method and benchmarked its performance against state-of-the-art methods across three threat models (refer to \Cref{settings}). As summarized in \Cref{tab_cifar10,tab_rn50,tab_vit,tab_vgg}, GreedyPixel consistently achieved higher ASRs than the benchmarks under the same conditions. White-box GreedyPixel attacks outperformed their black-box counterparts, which were constrained by a limited query budget. Additionally, GreedyPixel demonstrated greater effectiveness on a low-resolution dataset (CIFAR-10) than on a high-resolution one (ImageNet) in the black-box setting as only a subset of pixels could be modified within a restricted query budget, thereby limiting attack success. Furthermore, as shown in \Cref{tab_vit}, GreedyPixel maintained high ASRs and strong perceptual quality on ViT-Base-Patch16, confirming that the proposed approach is effective across both CNN- and Transformer-based models.

\begin{table}[!t]
    \centering
    \caption{Attack performance of WideResNet-28-10 \cite{croce2021robustbench} evaluated on CIFAR-10 \cite{krizhevsky2009learning}.}
    \label{tab_cifar10}
    \begin{threeparttable}
        \setlength{\tabcolsep}{0.7mm}{\begin{tabular}{cc|r@{\hspace{10pt}}|r@{\hspace{5pt}}|r@{\hspace{8pt}}|c|c}
            \toprule
            Threat Model&Attack&\multicolumn{1}{c|}{Time(s)$\downarrow$}&\multicolumn{1}{c|}{Avg. Q$\downarrow$}&\multicolumn{1}{c|}{ASR(\%)$\uparrow$}&LPIPS$\downarrow$&SSIM$\uparrow$\\
            \midrule
            \multirow{3}{*}{\begin{tabular}{c}White-box\\(Limited $\epsilon$)\end{tabular}}&APGD&1.7&\multicolumn{1}{c|}{\multirow{3}{*}{$\infty$}}&\textbf{100.0}&0.0006&0.981\\
            &AutoAttack&1.7&&\textbf{100.0}&0.0006&0.982\\
            &Ours&\textbf{1.3}&&\textbf{100.0}&\textbf{0.0001}&\textbf{0.994}\\
            \midrule
            \multirow{6}{*}{\begin{tabular}{c}Black-box\\(Limited $\epsilon$)\end{tabular}}&DeCoWA$^*$&28.1&\textbf{0}&8.7&0.0022&0.986\\
            &OPS&16.1&\textbf{0}&8.2&0.0028&0.990\\
            &GAA&3.0&\textbf{0}&12.2&0.0022&0.986\\
            &MCG&3.7&745&99.5&0.0013&0.978\\
            &SQBA$^*$&81.3&17,418&13.0&0.0016&0.988\\
            &Ours&\textbf{1.8}&1,997&\textbf{100.0}&\textbf{0.0002}&\textbf{0.995}\\
            \midrule
            \multirow{2}{*}{\begin{tabular}{c}Black-box\\(Unlimited $\epsilon$)\end{tabular}}&BruSLe&1.4&367&98.7&0.0168&0.876\\
            &Ours&\textbf{0.3}&\textbf{318}&\textbf{100.0}&\textbf{0.0023}&\textbf{0.964}\\
            \bottomrule
        \end{tabular}}
        \begin{tablenotes}
            \item $^*$DeCoWA \cite{lin2024boosting} and SQBA \cite{park2024hard} exhibited limited transferability when there was a significant discrepancy between the surrogate model (PreActResNet-18 \cite{wong2020fast}) and the target model (WideResNet-28-10 \cite{croce2021robustbench}). To ensure the correctness of our implementation, we evaluated their performance in a white-box attack setting (where the surrogate and target models are identical), achieving success rates of 70.5\% and 84.3\%, respectively. \textbf{Bold} indicates superior performance. 
        \end{tablenotes}
    \end{threeparttable}
\end{table}

\begin{table}[!t]
    \centering
    \caption{Attack performance of ResNet-50 \cite{croce2021robustbench} evaluated on ImageNet \cite{nips2017}.}
    \label{tab_rn50}
    \begin{threeparttable}
        \setlength{\tabcolsep}{0.7mm}{\begin{tabular}{cc|r@{\hspace{8pt}}|r@{\hspace{5pt}}|r@{\hspace{8pt}}|c|c}
            \toprule
            Threat Model&Attack&\multicolumn{1}{c|}{Time(s)$\downarrow$}&\multicolumn{1}{c|}{Avg. Q$\downarrow$}&\multicolumn{1}{c|}{ASR(\%)$\uparrow$}&LPIPS$\downarrow$&SSIM$\uparrow$\\
            \midrule
            \multirow{3}{*}{\begin{tabular}{c}White-box\\(Limited $\epsilon$)\end{tabular}}&APGD&\textbf{2.5}&\multicolumn{1}{c|}{\multirow{3}{*}{$\infty$}}&99.9&0.0351&0.954\\
            &AutoAttack&2.6&&\textbf{100.0}&0.0343&0.955\\
            &Ours&23.2&&\textbf{100.0}&\textbf{0.0006}&\textbf{0.997}\\
            \midrule
            \multirow{6}{*}{\begin{tabular}{c}Black-box\\(Limited $\epsilon$)\end{tabular}}&DeCoWA$^*$&536.3&\textbf{0}&8.0&0.0356&0.944\\
            &OPS&60.1&\textbf{0}&16.1&0.0233&0.976\\
            &GAA&\textbf{11.5}&\textbf{0}&19.5&0.0284&0.966\\
            &GFCS&259.1&12,217&42.2&0.0229&0.978\\
            &SQBA$^*$&101.4&15,538&21.6&0.0224&0.971\\
            &Ours&26.1&13,152&\textbf{60.0}&\textbf{0.0002}&\textbf{$\approx$1}\\
            \midrule
            \multirow{2}{*}{\begin{tabular}{c}Black-box\\(Unlimited $\epsilon$)\end{tabular}}&BruSLe&7.2&\textbf{913}&98.0&0.0616&0.977\\
            &Ours&\textbf{2.0}&980&\textbf{99.9}&\textbf{0.0232}&\textbf{0.985}\\
            \bottomrule
        \end{tabular}}
        \begin{tablenotes}
            \item $^*$DeCoWA \cite{lin2024boosting} and SQBA \cite{park2024hard} exhibited limited transferability when there was a significant discrepancy between the surrogate model (ConvNeXt \cite{singh2023revisiting}) and the target model (ResNet-50 \cite{croce2021robustbench}). To ensure the correctness of our implementation, we evaluated their performance in a white-box setting (where the surrogate and target models are identical), achieving success rates of 94.9\% and 99.6\%, respectively. \textbf{Bold} indicates superior performance. 
        \end{tablenotes}
    \end{threeparttable}
\end{table}

\begin{table}[!t]
    \centering
    \caption{Attack performance of ViT-Base-Patch16 (Vision Transformer) \cite{wu2020visual} evaluated on ImageNet \cite{nips2017}.}
    \label{tab_vit}
    \begin{threeparttable}
        \setlength{\tabcolsep}{0.7mm}{\begin{tabular}{cc|r@{\hspace{8pt}}|r@{\hspace{5pt}}|r@{\hspace{8pt}}|c|c}
            \toprule
            Threat Model&Attack&\multicolumn{1}{c|}{Time(s)$\downarrow$}&\multicolumn{1}{c|}{Avg. Q$\downarrow$}&\multicolumn{1}{c|}{ASR(\%)$\uparrow$}&LPIPS$\downarrow$&SSIM$\uparrow$\\
            \midrule
            \multirow{3}{*}{\begin{tabular}{c}White-box\\(Limited $\epsilon$)\end{tabular}}&APGD&\textbf{2.6}&\multicolumn{1}{c|}{\multirow{3}{*}{$\infty$}}&\textbf{100.0}&0.0349&0.958\\
            &AutoAttack&\textbf{2.6}&&\textbf{100.0}&0.0333&0.959\\
            &Ours&63.4&&\textbf{100.0}&\textbf{0.0013}&\textbf{0.995}\\
            \midrule
            \multirow{6}{*}{\begin{tabular}{c}Black-box\\(Limited $\epsilon$)\end{tabular}}&DeCoWA$^*$&536.3&\textbf{0}&3.8&0.0356&0.944\\
            &OPS&60.1&\textbf{0}&6.4&0.0233&0.976\\
            &GAA&\textbf{11.5}&\textbf{0}&9.0&0.0284&0.966\\
            &GFCS&347.3&15,769&24.0&0.0239&0.978\\
            &SQBA$^*$&112.3&18,024&9.6&0.0240&0.951\\
            &Ours&65.0&16,096&\textbf{37.8}&\textbf{0.0004}&\textbf{0.999}\\
            \midrule
            \multirow{2}{*}{\begin{tabular}{c}Black-box\\(Unlimited $\epsilon$)\end{tabular}}&BruSLe&10.4&\textbf{1,509}&98.6&0.0641&0.977\\
            &Ours&\textbf{6.6}&1,620&\textbf{100.0}&\textbf{0.0341}&\textbf{0.978}\\
            \bottomrule
        \end{tabular}}
        \begin{tablenotes}
            \item $^*$DeCoWA \cite{lin2024boosting} and SQBA \cite{park2024hard} exhibited limited transferability when there was a significant discrepancy between the surrogate model (ConvNeXt \cite{singh2023revisiting}) and the target model (ViT-Base-Patch16 \cite{wu2020visual}). To ensure the correctness of our implementation, we evaluated their performance in a white-box attack setting (where the surrogate and target models are identical), achieving success rates of 78\% and 83.1\%, respectively. \textbf{Bold} indicates superior performance. 
        \end{tablenotes}
    \end{threeparttable}
\end{table}

\begin{table}[!t]
    \centering
    \caption{Attack performance of VGG19-BN \cite{simonyan2015very} evaluated on ImageNet \cite{nips2017}.}
    \label{tab_vgg}
    \begin{threeparttable}
        \setlength{\tabcolsep}{0.7mm}{\begin{tabular}{cc|r@{\hspace{8pt}}|r@{\hspace{5pt}}|r@{\hspace{8pt}}|c|c}
            \toprule
            Threat Model&Attack&\multicolumn{1}{c|}{Time(s)$\downarrow$}&\multicolumn{1}{c|}{Avg. Q$\downarrow$}&\multicolumn{1}{c|}{ASR(\%)$\uparrow$}&LPIPS$\downarrow$&SSIM$\uparrow$\\
            \midrule
            \multirow{3}{*}{\begin{tabular}{c}White-box\\(Limited $\epsilon$)\end{tabular}}&APGD&\textbf{0.8}&\multicolumn{1}{c|}{\multirow{3}{*}{$\infty$}}&\textbf{100.0}&0.0254&0.964\\
            &AutoAttack&0.9&&\textbf{100.0}&0.0246&0.965\\
            &Ours&26.2&&\textbf{100.0}&\textbf{0.0012}&\textbf{0.996}\\
            \midrule
            \multirow{6}{*}{\begin{tabular}{c}Black-box\\(Limited $\epsilon$)\end{tabular}}&DeCoWA$^*$&536.3&\textbf{0}&36.6&0.0356&0.944\\
            &OPS&60.1&\textbf{0}&48.4&0.0233&0.976\\
            &GAA&\textbf{11.5}&\textbf{0}&47.6&0.0284&0.966\\
            &GFCS&125.0&6,198&71.5&0.0124&0.984\\
            &SQBA$^*$&39.4&9,846&50.1&0.0162&0.978\\
            &Ours&13.2&6,662&\textbf{84.7}&\textbf{0.0001}&\textbf{$\approx$1}\\
            \midrule
            \multirow{2}{*}{\begin{tabular}{c}Black-box\\(Unlimited $\epsilon$)\end{tabular}}&BruSLe&1.9&\textbf{391}&99.3&0.0580&0.978\\
            &Ours&\textbf{1.0}&477&\textbf{99.9}&\textbf{0.0111}&\textbf{0.993}\\
            \bottomrule
        \end{tabular}}
        \begin{tablenotes}
            \item $^*$DeCoWA \cite{lin2024boosting} and SQBA \cite{park2024hard} exhibited limited transferability when there was a significant discrepancy between the surrogate model (ConvNeXt \cite{singh2023revisiting}) and the target model (VGG19-BN \cite{simonyan2015very}). To ensure the correctness of our implementation, we evaluated their performance in a white-box setting (where the surrogate and target models are identical), achieving success rates of 99.2\% and 99.9\%, respectively. \textbf{Bold} indicates superior performance. 
        \end{tablenotes}
    \end{threeparttable}
\end{table}

\subsection{Attack Efficiency}
We assessed the efficiency of the GreedyPixel attack method from two perspectives: execution time and query efficiency. As shown in \Cref{tab_cifar10,tab_rn50,tab_vit,tab_vgg}, GreedyPixel outperformed all black-box attacks in execution time, making it the fastest method. Although it required more queries than query-based attacks such as DeCoWA \cite{lin2024boosting}, GFCS \cite{lord2022attacking}, and MCG \cite{yin2023generalizable}, it achieved superior ASRs, lower execution times, and better visual quality (higher SSIM, lower LPIPS). This is due to GreedyPixel's ability to compute eight queries in parallel ($\{-\epsilon,+\epsilon\}^3=8$), whereas other attacks query sequentially. Additionally, when the $L_\infty$-norm constraint ($\epsilon$) was relaxed, GreedyPixel surpassed the attack efficiency of a state-of-the-art sparse attack, BruSLe \cite{vo2024brusleattack}.

The complexity of attacking high-resolution images (ImageNet) increases more sharply for GreedyPixel than for low-resolution images (CIFAR-10). Despite this, GreedyPixel achieved a 100\% ASR for white-box attacks on CIFAR-10, outperforming methods such as APGD \cite{croce2020reliable} and AutoAttack \cite{croce2020reliable} in execution time. However, when applied to ImageNet, GreedyPixel remained effective but was less efficient in execution time than APGD and AutoAttack, reflecting the trade-off between resolution and computational cost.

\begin{figure*}[!t]
    \centering
    \includegraphics[width=\linewidth]{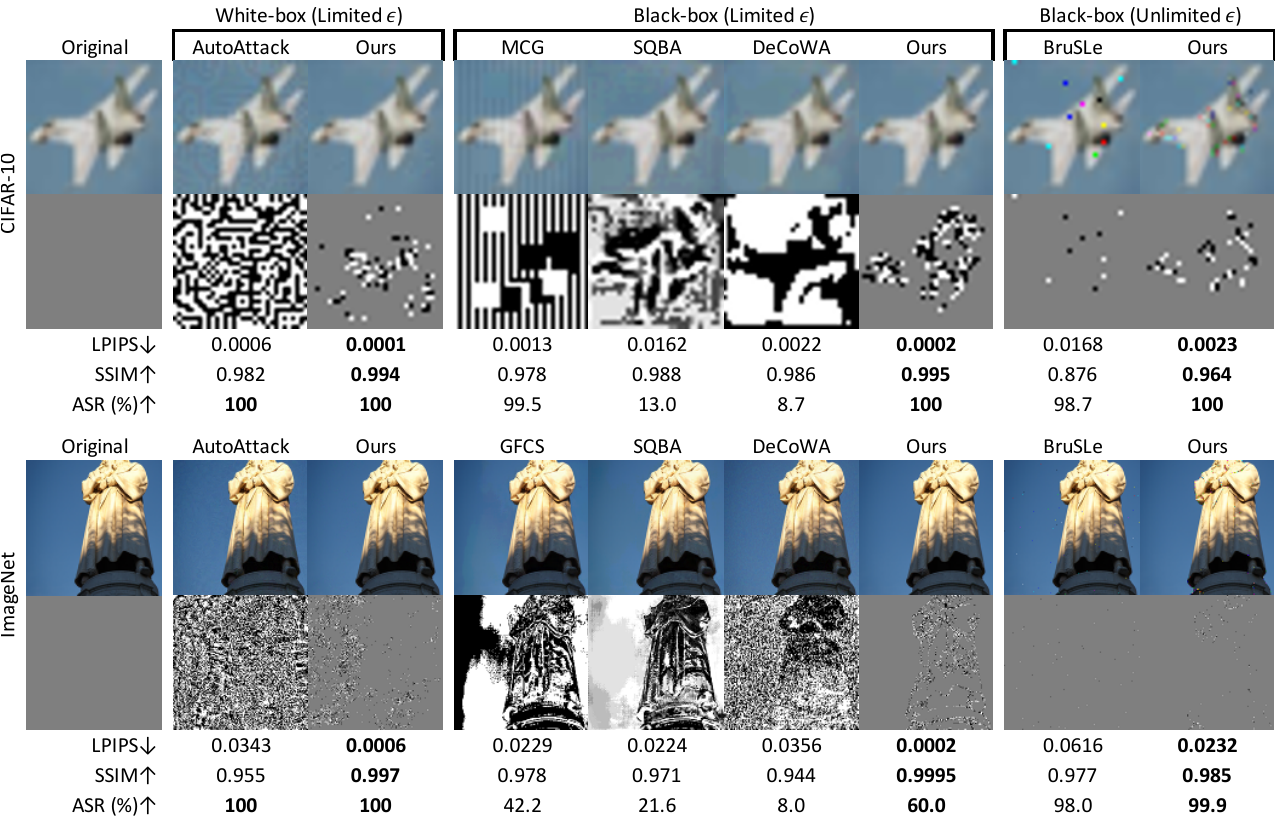}
    \caption{Comparison of adversarial examples generated by different attack algorithms in terms of visual quality and perturbation appearance. The proposed GreedyPixel method achieved the highest visual quality by selectively modifying only critical pixels. The perturbations were normalized for visualization using \Cref{eq_normalization}, with \textbf{bold} indicating superior performance. Notably, while the BruSLe attack \cite{vo2024brusleattack} introduced fewer perturbed pixels than GreedyPixel, the visibility of these perturbations was higher due to their spatial placement.}
    \label{fig_qualitative}
\end{figure*}

\subsection{Visual Quality}
We computed the SSIM and LPIPS scores between the original inputs and adversarial examples to assess the visibility of adversarial perturbations. Additionally, we visualized these perturbations to compare the qualitative differences among algorithms, highlighting variations in visual quality. The grayscale perturbation illustration $\delta_{Gray}$ in \Cref{fig_qualitative} was generated in two steps: first, the perturbations $\delta$ were normalized from the range $[\epsilon_{min},\epsilon_{max}]$ to $[0,1]$ to produce $\delta_{RGB}$, and then $\delta_{RGB}$ was converted to grayscale using the NTSC formula \cite{livingston1954colorimetric}:
\begin{equation}
    \begin{aligned}
        \delta_{RGB}=&\frac{\delta-\epsilon_{min}}{\epsilon_{max}-\epsilon_{min}},\\
        \delta_{Gray}=&0.299\cdot\delta_{RGB}.R+\\
        &\ 0.587\cdot\delta_{RGB}.G+0.114\cdot\delta_{RGB}.B,
    \end{aligned}
    \label{eq_normalization}
\end{equation}
where $\delta_{RGB}.R$, $\delta_{RGB}.G$, and $\delta_{RGB}.B$ represent the red, green, and blue channels, respectively.

As shown in \Cref{tab_cifar10,tab_rn50,tab_vit,tab_vgg}, GreedyPixel consistently outperformed the other attacks in visual quality, with higher SSIM and lower LPIPS scores. This is further demonstrated in \Cref{fig_qualitative}, where our method introduced less noticeable noise, especially in black-box settings. Similar to the BruSLe attack \cite{vo2024brusleattack}, GreedyPixel modified fewer pixels than methods such as AutoAttack \cite{croce2020reliable}, DeCoWA \cite{lin2024boosting}, GFCS \cite{lord2022attacking}, MCG \cite{yin2023generalizable}, and SQBA \cite{park2024hard}, leading to more subtle and imperceptible changes. By focusing perturbations on robust features such as complex objects and edges, rather than the background, GreedyPixel achieved superior visual quality, making its perturbations minimally invasive yet highly effective.

\begin{table}[!t]
    \centering
    \caption{User study on visual quality.}
    \label{tab_survey}
    \begin{threeparttable}
        \setlength{\tabcolsep}{1mm}{\begin{tabular}{cc|r@{\hspace{9pt}}r@{\hspace{10pt}}|r@{\hspace{9pt}}r@{\hspace{10pt}}}
            \toprule
            \multirow{2}{*}{Threat Model}&\multirow{2}{*}{Attack}&\multicolumn{2}{c|}{CIFAR-10}&\multicolumn{2}{c}{ImageNet}\\
            &&\multicolumn{1}{c}{ASR(\%)$\uparrow$}&\multicolumn{1}{c|}{Invisible$\uparrow$}&\multicolumn{1}{c}{ASR(\%)$\uparrow$}&\multicolumn{1}{c}{Invisible$\uparrow$}\\
            \midrule
            \multirow{2}{*}{\begin{tabular}{c}White-box\\(Limited $\epsilon$)\end{tabular}}&AutoAttack&\textbf{100.0}&32/50&\textbf{100.0}&38/50\\
            &Ours&\textbf{100.0}&\textbf{46/50}&\textbf{100.0}&\textbf{49/50}\\
            \midrule
            \multirow{3}{*}{\begin{tabular}{c}Black-box\\(Limited $\epsilon$)\end{tabular}}&MCG&99.5&8/50&-&-\\
            &GFCS&-&-&42.2&43/50\\
            &Ours&\textbf{100.0}&\textbf{47/50}&\textbf{60.0}&\textbf{49/50}\\
            \midrule
            \multirow{2}{*}{\begin{tabular}{c}Black-box\\(Unlimited $\epsilon$)\end{tabular}}&BruSLe&98.7&0/50&98.0&2/50\\
            &Ours&\textbf{100.0}&\textbf{4/50}&\textbf{99.9}&\textbf{20/50}\\
            \bottomrule
        \end{tabular}}
        \begin{tablenotes}
            \item ``Invisible" denotes percentage of examples with imperceptible perturbations. \textbf{Bold} indicates superior performance. MCG and GFCS did not release their code for ImageNet and CIFAR-10 datasets, respectively.
        \end{tablenotes}
    \end{threeparttable}
\end{table}

\begin{table*}[!t]
    \centering
    \caption{Attack Performance (\%) against adversarial defenses in counterattack experiments.}
    \label{tab_defense}
    \begin{threeparttable}
        \setlength{\tabcolsep}{2.7mm}{\begin{tabular}{ccr@{\hspace{22pt}}r@{\hspace{33pt}}r@{\hspace{17pt}}r@{\hspace{15pt}}r@{\hspace{17pt}}r@{\hspace{32pt}}}
            \toprule
            Threat Model&Attack&\multicolumn{1}{c}{No Defense $\uparrow$}&\multicolumn{1}{c}{Purification$^a$ \cite{nie2022diffusion} $\uparrow$}&\multicolumn{1}{c}{AT \cite{wang2023better} $\uparrow$}&\multicolumn{1}{c}{AT \cite{bartoldson2024adversarial} $\uparrow$}&\multicolumn{1}{c}{AT \cite{debenedetti2023light} $\uparrow$}&\multicolumn{1}{c}{Detection$^b$ \cite{meng2017magnet}  $\downarrow$}\\
            \midrule
            \multirow{3}{*}{\begin{tabular}{c}White-box\\(Limited $\epsilon$)\end{tabular}}&APGD&\textbf{100.0}&20.5&16.0&13.1&20.7&10.7\\
            &AutoAttack&\textbf{100.0}&21.5&\textbf{16.7}&\textbf{14.2}&\textbf{21.8}&10.6\\
            &GreedyPixel (ours)&\textbf{100.0}&\textbf{25.4}&16.5&14.0&21.4&\textbf{8.4}\\
            \midrule
            \multirow{3}{*}{\begin{tabular}{c}Black-box\\(Limited $\epsilon$)\end{tabular}}&DeCoWA$^c$&8.7&16.2&9.9&9.0&13.4&\textbf{7.7}\\
            &MCG&99.5&13.2&15.3&13.3&20.0&7.9\\
            &SQBA&13.0&17.5&10.3&9.3&15.1&8.2\\
            &GreedyPixel (ours)&\textbf{100.0}&\textbf{23.4}&\textbf{16.4}&\textbf{14.0}&\textbf{21.3}&8.5\\
            \midrule
            \multirow{2}{*}{\begin{tabular}{c}Black-box\\(Unlimited $\epsilon$)\end{tabular}}&BruSLe&98.7&30.8&72.5&75.5&57.1&81.9\\
            &GreedyPixel (ours)&\textbf{100.0}&\textbf{36.2}&\textbf{84.6}&\textbf{82.7}&\textbf{93.2}&\textbf{31.5}\\
            \bottomrule
        \end{tabular}}
        \begin{tablenotes}
            \item $^a$Diffusion timestep=0.1.
            $^b$The detector was trained on adversarial examples generated by Adversarial Patch \cite{brown2017adversarial}, DeepFool \cite{moosavi2016deepfool}, FGSM \cite{goodfellow2014explaining}, PGD \cite{madry2018towards}, APGD \cite{croce2020reliable}, Spatial Transformation \cite{xiao2018spatially}, and Square \cite{andriushchenko2020square} attacks. The threshold was set for a 5\% clean rejection rate.
            $^c$DeCoWA \cite{lin2024boosting} demonstrated suboptimal performance when using the surrogate model \cite{wong2020fast} irrespective of whether it targeted non-robust or robust models. The higher ASR against DiffPure is attributable to distortion dominating over adversarial manipulations.
            \textbf{Bold} indicates superior performance.
        \end{tablenotes}
    \end{threeparttable}
\end{table*}

\subsection{User Study on Visual Quality}
We conducted a user study to evaluate the human perceptual quality of adversarial examples. We randomly selected 100 original images and generated adversarial counterparts using seven methods: white-box AutoAttack \cite{croce2020reliable}, black-box GFCS \cite{lord2022attacking} and MCG \cite{yin2023generalizable}, sparse BruSLe \cite{vo2024brusleattack}, and GreedyPixel, under three threat models. The attacks were applied to the CIFAR-10 (low-resolution) and ImageNet (high-resolution) datasets. Thirty participants with varying vision levels were asked to identify adversarial examples that were indistinguishable from the originals. The samples were presented in random order to minimize bias.

As shown in \Cref{tab_survey}, GreedyPixel achieved superior visual quality, with more adversarial examples judged indistinguishable from the originals, especially in white-box and black-box settings under limited $\epsilon$. Under unlimited $\epsilon$ conditions, both GreedyPixel and BruSLe perturbations became more detectable, though GreedyPixel still outperformed BruSLe. Additionally, perturbations on CIFAR-10 were more easily detected than those on ImageNet, suggesting that high-resolution images present a greater challenge for defense mechanisms.

\subsection{Effectiveness Against Adversarial Defenses}
To assess GreedyPixel's effectiveness under realistic deployment scenarios, we evaluated it against multiple representative defenses, including a state-of-the-art purification method (DiffPure~\cite{nie2022diffusion}), three AT models~\cite{wang2023better,bartoldson2024adversarial,debenedetti2023light} (including the top-ranked robust model~\cite{bartoldson2024adversarial} from RobustBench~\cite{croce2021robustbench}), and an adversarial detector~\cite{meng2017magnet}, using the CIFAR-10 dataset. The adversarial examples were generated using 20,000 queries without early stopping. As shown in \Cref{tab_defense}, GreedyPixel outperformed the other methods, achieving comparable or higher ASRs and similar or lower adversarial detection rates across all three threat models. This demonstrates GreedyPixel's enhanced effectiveness and adaptability in bypassing adversarial defenses.

\begin{figure*}[!t]
    \centering
    \includegraphics[width=6in]{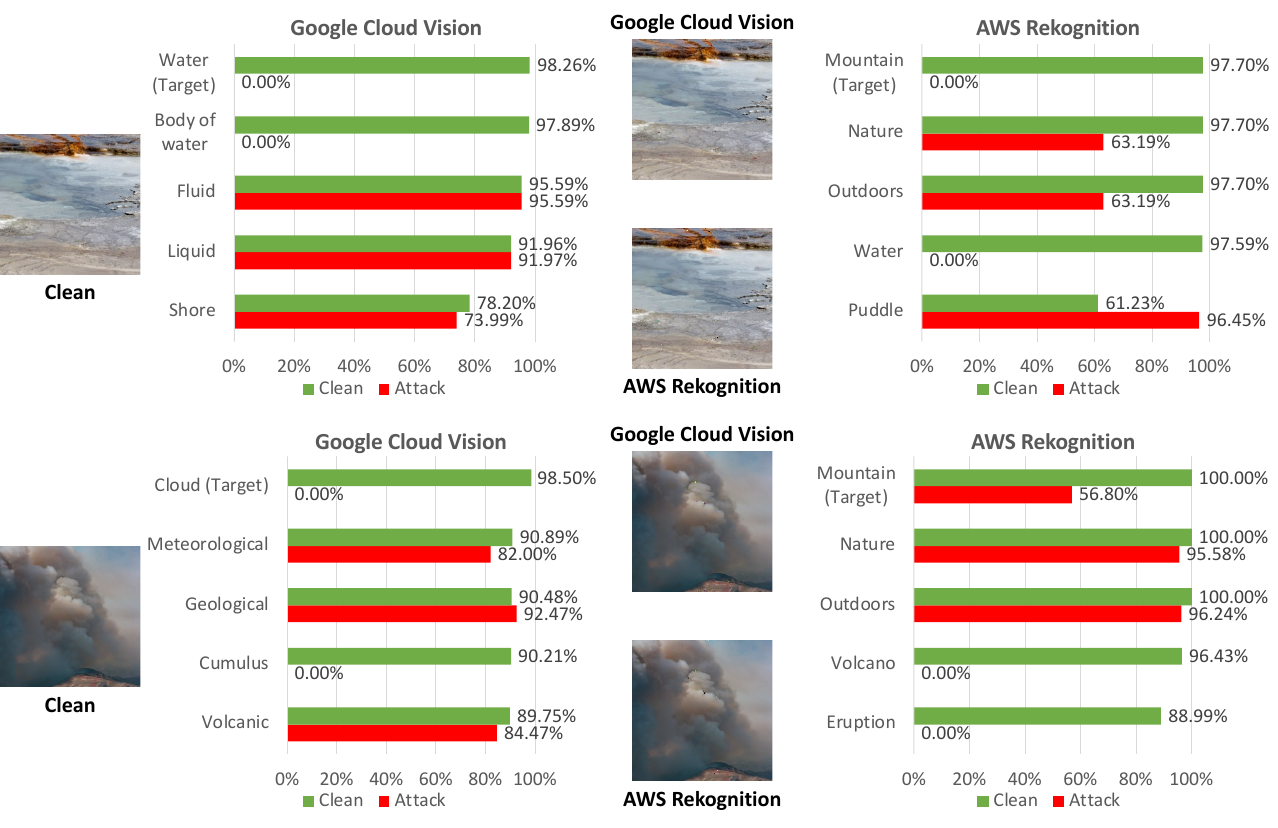}
    \caption{Results of GreedyPixel attacks on commercial online vision APIs. Within a strict budget of 400 queries (perturbing fewer than 0.1\% of pixels, \ie $\leq 50$ pixels under unlimited $\epsilon$), the confidence scores of the ground-truth class were significantly reduced—sometimes to zero. These results demonstrate that GreedyPixel remains effective even in real-world, production scenarios.}
    \label{fig_apis}
\end{figure*}

\subsection{Attacks Against Commercial Online Vision APIs}
To further evaluate the real-world applicability of GreedyPixel, we conducted black-box attacks against two widely deployed commercial systems: Google Cloud Vision~\cite{googlecloudvision} and AWS Rekognition~\cite{awsrekognition}. We used a subset of ImageNet samples as inputs and restricted the maximum query budget to 400, resulting in extremely sparse perturbations ($<0.1\%$ pixels; \ie fewer than 50 pixels modified under unlimited $\epsilon$). 

As shown in \Cref{fig_apis}, GreedyPixel consistently reduced the confidence score of the correct class (sometimes to zero) within the tight budget. The qualitative results demonstrate that GreedyPixel generalizes beyond academic benchmarks and is highly effective in attacking real-world API-based vision models despite the limited number of queries and the practical constraints of commercial services. This highlights its practical relevance for assessing the robustness of deployed systems.

\section{Ablation Study}
\label{ablation}

\begin{figure*}[!t]
    \centering
    \includegraphics[width=\linewidth]{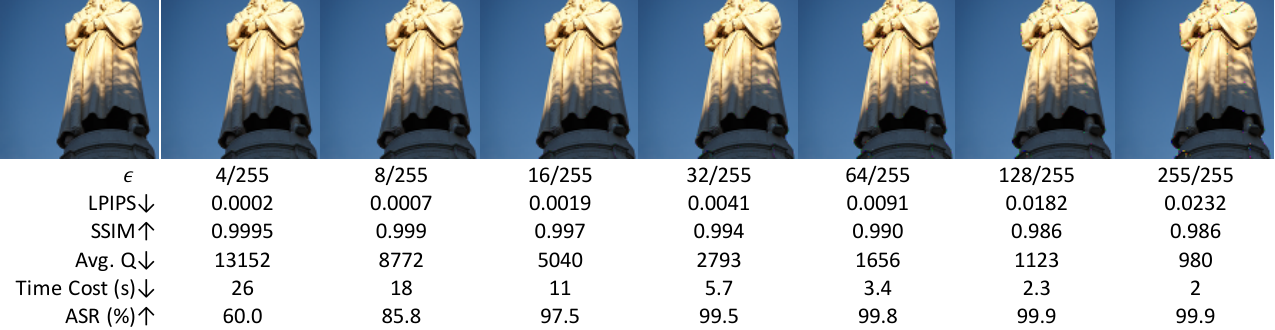}
    \caption{Visual quality, efficiency, and ASRs under different $L_\infty$-norm attack budgets. GreedyPixel supports larger attack budgets, trading off slight degradation in visual quality (as measured by SSIM and LPIPS) for improved efficiency and higher ASRs.}
    \label{fig_diffeps}
\end{figure*}

\subsection{Effect of Attack Constraint}
We investigated the effect of increasing attack budgets. The visual quality comparisons in \Cref{fig_qualitative} show that with a small $L_\infty$-norm constraint of 4/255, our method substantially outperformed others in SSIM and LPIPS scores. This indicates that increasing the attack budget can maintain superior visual quality while improving ASRs and attack efficiency. \Cref{fig_diffeps} further supports this, showing that slight degradation in visual quality (as reflected in SSIM and LPIPS) can lead to improved attack efficiency and higher ASRs.

\subsection{Maximum Query Limit}
\begin{figure*}[!t]
    \centering
    \includegraphics[width=\linewidth]{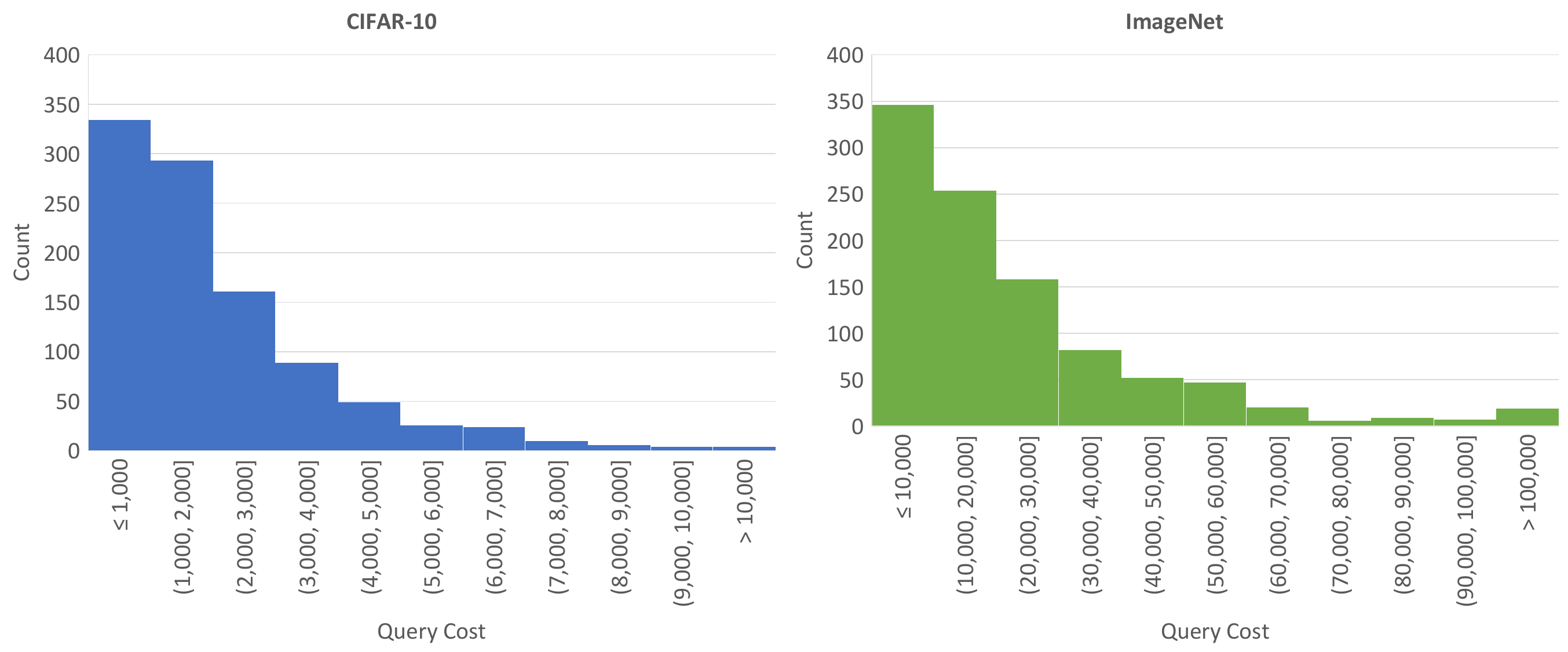}
    \caption{Query costs and distribution of successful adversarial examples. This figure illustrates the distribution of query costs among successful adversarial examples, demonstrating that the majority of these examples require relatively few queries to succeed. While increasing the allowed number of queries can improve attack performance, the rate of improvement diminishes as the query limit grows larger. Beyond a certain point (10,000 queries / 1,250 pixels for CIFAR-10, or 100,000 queries / 12,500 pixels for ImageNet), further increases mainly affect the few remaining unsuccessful cases, leading to marginal gains in overall performance.}
    \label{fig_querycost}
\end{figure*}

In query-based adversarial attacks, the maximum query limit markedly affects attack effectiveness, especially for challenging images. This limit is tied to the number of manipulable pixels, as described in \Cref{eq_onepass}. The results of our analysis, summarized in \Cref{tab_query}, illustrate the effects of varying the query limit across performance metrics. For example, setting a low query threshold (\eg~5,000 queries for CIFAR-10, equating to 625 pixels out of 1,024, or 20,000 queries for ImageNet, affecting 2,500 pixels out of 50,176) limits the number of pixels that can be altered, resulting in lower ASRs. However, increasing the threshold beyond 10,000 queries (1,250 pixels) for CIFAR-10 or 100,000 queries (12,500 pixels) for ImageNet leads to minimal improvements in effectiveness. In these higher-query settings, metrics such as Avg. Q, execution time, and perceptual measures like LPIPS and SSIM show little variation. This plateau effect occurs because the extra queries mainly affect previously unsuccessful cases, as seen in \Cref{fig_querycost}. Moreover, ASRs in higher-query settings indicate that, with sufficient time and resources, GreedyPixel can achieve results comparable to gradient-based methods without requiring access to gradient information, as discussed in \Cref{greedy}.

\begin{table}[!t]
    \centering
    \caption{Black-box attack performance evaluation at a fixed perturbation limit of $\epsilon = 4/255$, under various maximum query constraints.}
    \label{tab_query}
    \begin{threeparttable}
        \setlength{\tabcolsep}{0.6mm}{\begin{tabular}{cr@{\hspace{5pt}}|r@{\hspace{5pt}}|r@{\hspace{12pt}}|r@{\hspace{11pt}}|c|c}
            \toprule
            Resolution&\multicolumn{1}{c|}{Max. Q}&\multicolumn{1}{c|}{Avg. Q $\downarrow$}&\multicolumn{1}{c|}{Time (s) $\downarrow$}&\multicolumn{1}{c|}{ASR (\%) $\uparrow$}&LPIPS $\downarrow$&SSIM $\uparrow$\\
            \midrule
            \multirow{4}{*}{\begin{tabular}{c}$32 \times 32$\\(CIFAR-10)\end{tabular}}&5K&1,863&1.7&92.6&0.0002&0.996\\
            &10K&1,993&1.8&99.6&0.0002&0.995\\
            &15K&1,997&1.8&100.0&0.0002&0.995\\
            &20K&1,997&1.8&100.0&0.0002&0.995\\
            \midrule
            \multirow{4}{*}{\begin{tabular}{c}$64 \times 64$\\(ImageNet)\end{tabular}}&10K&2,687&4.7&94.9&0.0003&0.999\\
            &20K&2,957&5.1&98.8&0.0003&0.999\\
            &50K&3,108&5.5&99.7&0.0003&0.998\\
            &100K&3,259&5.5&100.0&0.0003&0.998\\
            \midrule
            \multirow{5}{*}{\begin{tabular}{c}$224 \times 224$\\(ImageNet)\end{tabular}}&20K&13,152&26.4&60.0&0.0002&$\approx$1\\
            &50K&19,598&40.0&89.2&0.0004&0.999\\
            &100K&21,816&44.4&98.1&0.0005&0.999\\
            &200K&22,617&47.0&99.6&0.0005&0.999\\
            &400K&23,003&47.1&99.9&0.0005&0.999\\
            \bottomrule
        \end{tabular}}
    \end{threeparttable}
\end{table}

\subsection{Image Resolution}
As shown in \Cref{tab_query}, under an unconstrained query setting, ASRs approach 100\% as image resolution increases, resulting in outcomes similar to those for low-resolution images, but at higher computational cost. However, under a constrained query budget, higher-resolution images lead to reduced ASRs, highlighting the increased difficulty GreedyPixel faces in attacking high-resolution images. These results demonstrate the importance of an appropriate query budget that reflects the resolution of the images.

\begin{table}[!t]
    \centering
    \caption{Performance of priority map updates across different periods, using WideResNet-28-10 \cite{croce2021robustbench} on CIFAR-10 \cite{krizhevsky2009learning}.}
    \label{tab_period}
    \begin{threeparttable}
        \setlength{\tabcolsep}{1.2mm}{\begin{tabular}{c|cc|r@{\hspace{13pt}}|c|c}
            \toprule
            Threat Model&Sparsity&Update Period&\multicolumn{1}{c|}{ASR (\%) $\uparrow$}&LPIPS $\downarrow$&SSIM $\uparrow$\\
            \midrule
            \multirow{7}{*}{\begin{tabular}{c}White-box\\(Limited $\epsilon$)\end{tabular}}&10\%&Disable&50.3&\textbf{$\approx$0}&\textbf{0.997}\\
            &30\%&Disable&90.3&\textbf{$\approx$0}&0.995\\
            &50\%&Disable&98.6&0.0001&0.994\\
            &Disable&$N=100$&97.4&\textbf{$\approx$0}&0.995\\
            &Disable&$N=300$&99.7&\textbf{$\approx$0}&0.994\\
            &Disable&$N=500$&\textbf{100.0}&0.0001&0.994\\
            &Disable&Disable&\textbf{100.0}&0.0001&0.994\\
            \midrule
            \multirow{7}{*}{\begin{tabular}{c}Black-box\\(Limited $\epsilon$)\end{tabular}}&10\%&Disable&29.9&\textbf{$\approx$0}&\textbf{0.998}\\
            &30\%&Disable&86.1&0.0001&0.997\\
            &50\%&Disable&97.6&0.0002&0.996\\
            &Disable&$N=100$&40.0&\textbf{$\approx$0}&\textbf{0.998}\\
            &Disable&$N=300$&91.6&0.0001&0.996\\
            &Disable&$N=500$&98.8&0.0002&0.996\\
            &Disable&Disable&\textbf{100.0}&0.0002&0.995\\
            \bottomrule
        \end{tabular}}
        \begin{tablenotes}
            \item \textbf{Bold} indicates superior performance. 
        \end{tablenotes}
    \end{threeparttable}
\end{table}

\subsection{Periodicity of Updating Priority Map}
\label{ablation_update}
We designed the priority map update frequency to improve visual quality of adversarial examples, as measured by LPIPS and SSIM, with a trade-off in ASR. By updating the map and resetting the adversary sequence to the most critical pixel, the attack was guided to focus on more important pixels. As shown in \Cref{tab_period}, increasing the update frequency (\ie~reducing $N$ in \Cref{alg_GreedyPixel}) improved LPIPS and SSIM scores, indicating better visual quality, but reduced ASR. Additionally, we explored an alternative strategy in which sparsity was fixed by generating the priority map once and resetting the adversary sequence every $N$ iterations. Such alternative improved visual quality but caused a substantial drop in ASR. Due to its lower practicality, we did not focus on this alternative.

\begin{table*}[!t]
    \centering
    \caption{Attack performance with a fixed perturbation limit of $\epsilon = 4/255$ and a maximum query budget of 20,000 (including white-box attacks), evaluated across various surrogate models.}
    \label{tab_surrogate}
    \begin{threeparttable}
        \setlength{\tabcolsep}{3.7mm}{\begin{tabular}{cccr@{\hspace{13pt}}r@{\hspace{18pt}}r@{\hspace{18pt}}cc}
            \toprule
            Dataset&Target Model&Surrogate Model&\multicolumn{1}{c}{Avg. Q $\downarrow$}&\multicolumn{1}{c}{Time (s) $\downarrow$}&\multicolumn{1}{c}{ASR (\%) $\uparrow$}&LPIPS $\downarrow$&SSIM $\uparrow$\\
            \midrule
            \multirow{5}{*}{CIFAR-10}&\multirow{5}{*}{WideResNet-28-10 \cite{croce2021robustbench}}&WideResNet-28-10 \cite{croce2021robustbench}$^a$&1,510&1.3&100.0&0.0001&0.994\\
            &&PreActResNet-18 \cite{wong2020fast}&1,997&1.8&100.0&0.0002&0.995\\
            &&PreActResNet-18 \cite{gowal2021improving}&2,052&1.8&100.0&0.0002&0.995\\
            &&RaWideResNet-70-16 \cite{peng2023robust}&2,040&1.8&100.0&0.0002&0.995\\
            &&Random$^b$&2,220&2.1&100.0&0.0001&0.994\\
            \midrule
            \multirow{5}{*}{ImageNet}&\multirow{5}{*}{ResNet-50 \cite{croce2021robustbench}}&ResNet-50 \cite{croce2021robustbench}$^a$&7,826&16.4&86.1&0.0005&0.997\\
            &&ConvNeXt \cite{singh2023revisiting}&13,152&26.1&60.0&0.0002&$\approx$1\\
            &&Swin \cite{liu2023comprehensive}&13,242&27.0&59.2&0.0003&0.999\\
            &&ResNet-18 \cite{salman2020adversarially}&13,563&27.1&57.6&0.0002&0.999\\
            &&Random$^b$&14,149&27.4&54.9&0.0004&0.997\\
            \bottomrule
        \end{tabular}}
        \begin{tablenotes}
            \item $^a$Denotes white-box attacks where surrogate model was identical to target model.
            $^b$Denotes use of random map with no priority calculation for pixel selection.
        \end{tablenotes}
    \end{threeparttable}
\end{table*}

\begin{figure}[!t]
    \centering
    \includegraphics[width=\linewidth]{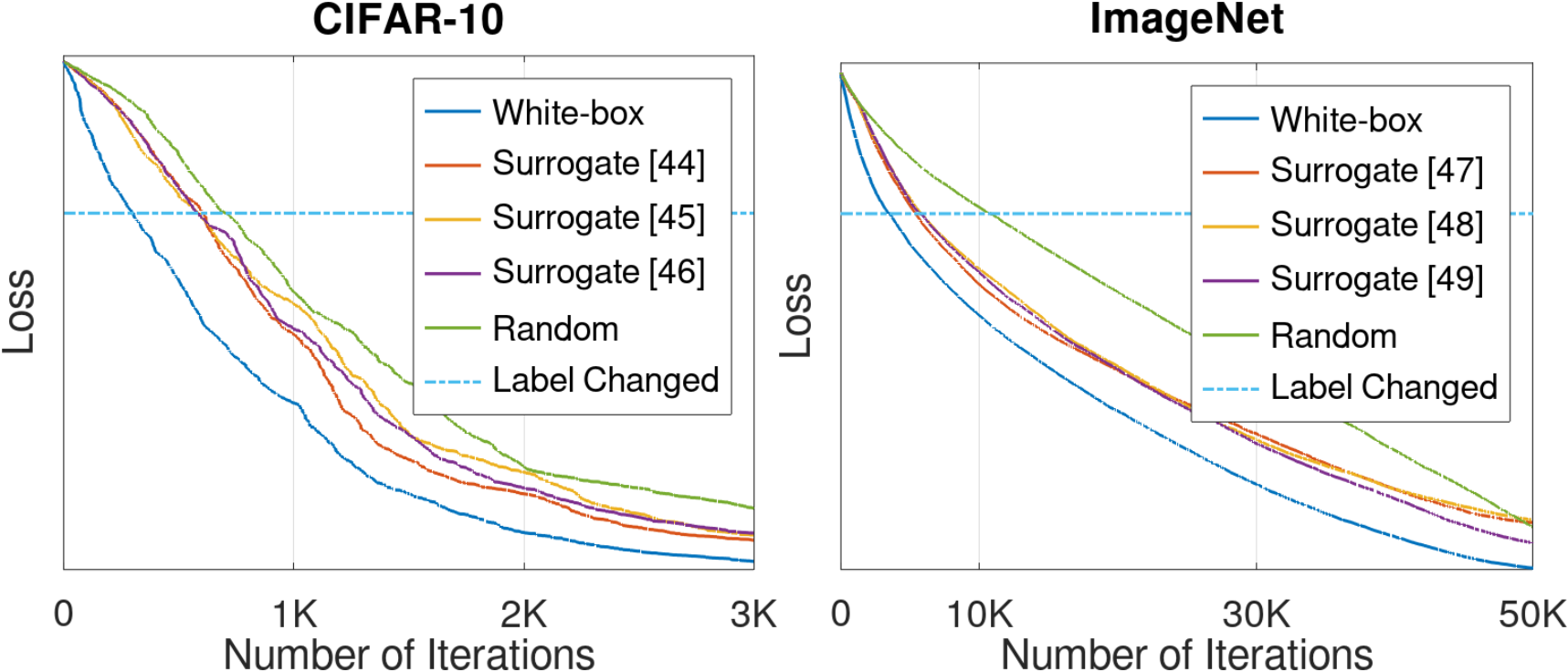}
    \caption{Query efficiency of proposed pixel-wise priority map under black-box and white-box settings. The proposed pixel-wise priority map effectively reduces the number of queries, with potential for further enhancement when incorporating white-box knowledge.}
    \label{fig_loss}
\end{figure}

\subsection{Priority Map and Surrogate Models}
\label{ablation_surrogate}
We investigated the effect of using priority maps learned from different surrogate models, including two PreActResNet-18 variants \cite{wong2020fast,gowal2021improving} and RaWideResNet-70-16 \cite{peng2023robust} for CIFAR-10, and ConvNeXt \cite{singh2023revisiting}, Swin \cite{liu2023comprehensive}, and ResNet-18 \cite{salman2020adversarially} for ImageNet, which together represent a diverse range of architectures and robustness levels.

As shown in \Cref{tab_surrogate}, using the priority map noticeably improves attack efficiency, reducing average query count and execution time compared with the ``Random" approach, especially under query constraints (\eg~for high-resolution images like those in ImageNet). ``White-box" knowledge further enhances both effectiveness and efficiency by leveraging precise gradient information. Additionally, adversarially trained models used as surrogates improve the visual quality compared with non-robust models, although the gains are modest.

These findings are illustrated in \Cref{fig_loss}, which shows that the priority map accelerates convergence by reducing the number of queries required to achieve loss values comparable to those with the ``Random" approach. The ``White-box" curves reflect superior performance, reinforcing the value of our strategy, particularly for high-resolution images, for which query efficiency is a critical factor.

\section{Discussion of Alternative Strategies}
We have demonstrated that the priority map is crucial for efficiency. Although GreedyPixel achieves strong performance, its query cost increases with image resolution, posing a computational challenge for very high-resolution inputs.

One potential approach is to simply increase the query budget until success is achieved. However, this is often impractical due to a long execution time, the risk of query blocking in deployed systems, and the resulting degradation of visual quality caused by excessive perturbations.

Another direction is to explore alternative priority maps, such as region-, objective-, edge-, or attention-based maps. While these approaches group pixels with the same priority, they require processing all pixels within each group, which typically reduces efficiency compared with per-pixel ordering. Our ablation study described in \Cref{ablation_update} demonstrated that re-ranking frequency plays an important role in balancing visual quality and attack success, suggesting that adaptive scheduling of priority refresh may further improve efficiency.

Given current knowledge, we believe that our algorithm represents an optimal compromise: it guarantees monotonic loss decrease, preserves high perceptual quality, and maintains query efficiency. Nevertheless, reducing query cost for very high-resolution images remains an open problem, and we plan to investigate advanced solutions such as hierarchical pixel selection and gradient-free attention guidance in future work.

\section{Conclusion and Future Research}
We have presented the strongest black-box adversarial attack to date, \textit{GreedyPixel}, which achieves robustness near that of white-box attacks without requiring gradient information. It is based on a novel strategy for black-box learning that supports fine-grained manipulations and provides precise control. By simulating white-box-like optimization while overcoming the local optima limitations of gradient descent, GreedyPixel bridges the gap between black-box and white-box learning. By leveraging a greedy algorithm and priority map, our approach makes brute-force adversarial learning feasible with a reasonable computational cost, achieving near-global optimality without gradient reliance. This advancement enhances both adversarial learning and general optimization techniques.

Future work will explore alternative approaches to constructing the priority map to further enhance the efficiency of greedy learning. Additionally, we will demonstrate the applicability of GreedyPixel for manipulating sensitive entities, such as the latent codes of diffusion models, to achieve precise image editing under black-box conditions.
{
    \bibliographystyle{ieeetr}
    \bibliography{main}
}
\begin{IEEEbiography}[{\includegraphics[width=1in,height=1.25in,clip,keepaspectratio]{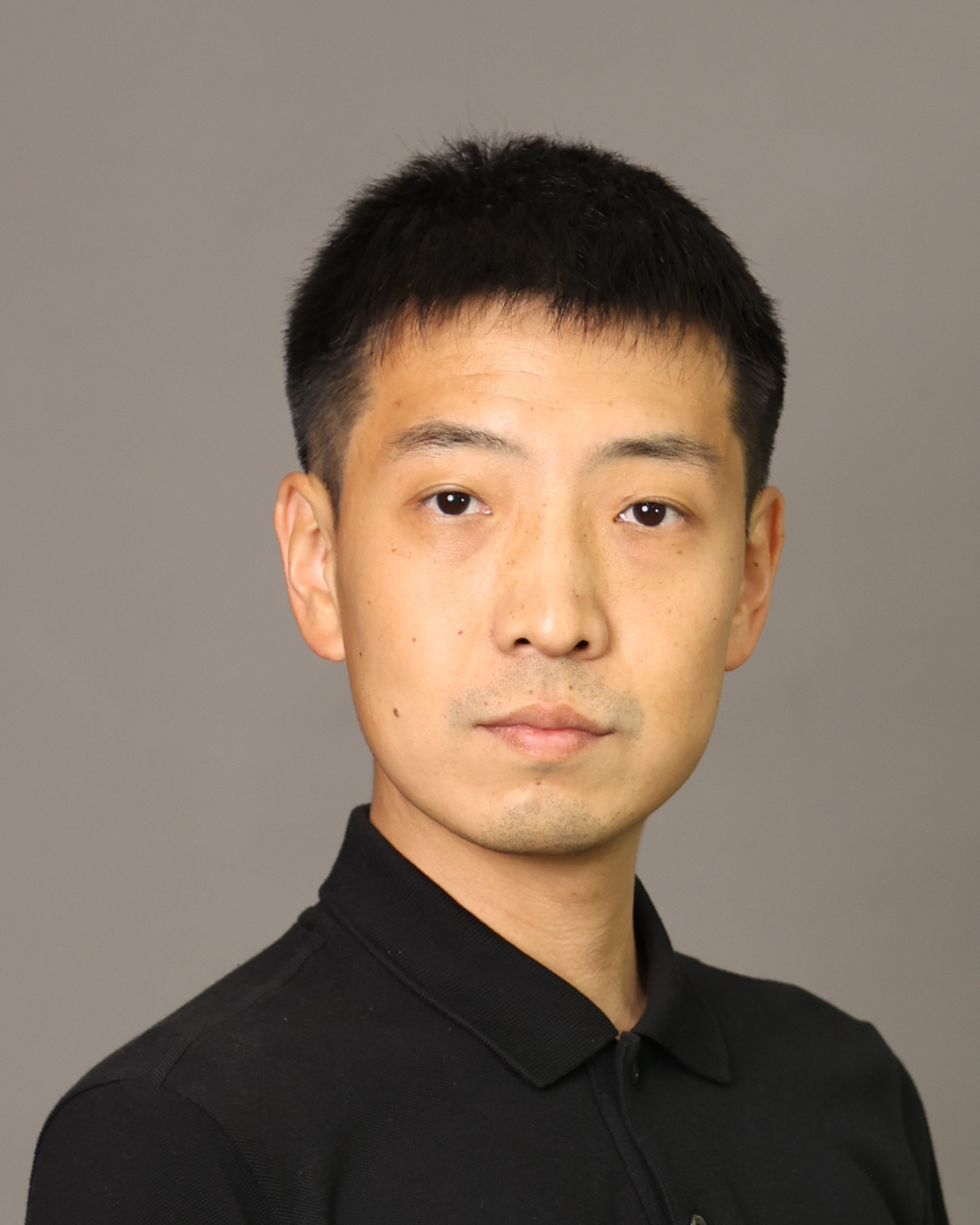}}]{Hanrui Wang}
received his B.S. degree in Electronic Information Engineering from Northeastern University (China) in 2011. He left the IT industry from a director position in 2019 to pursue a research career and received his Ph.D. in Computer Science from Monash University, Australia, in January 2024. He is currently working as a Postdoctoral Researcher with the Echizen Laboratory at the National Institute of Informatics (NII) in Tokyo, Japan. His research interests include AI security and privacy, particularly adversarial machine learning.
\end{IEEEbiography}

\begin{IEEEbiography}[{\includegraphics[width=1in,height=1.25in,clip,keepaspectratio]{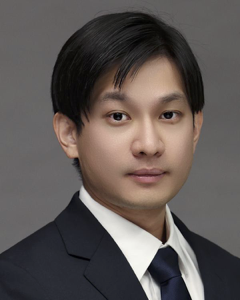}}]{Ching-Chun Chang}
received the PhD in Computer Science from the University of Warwick, UK, in 2019. He is currently a Project Assistant Professor with the National Institute of Informatics, Japan. He participated in a short-term scientific mission supported by European Cooperation in Science and Technology Actions at the Faculty of Computer Science, Otto von Guericke University of Magdeburg, Germany, in 2016. He was granted the Marie-Curie fellowship and participated in a research and innovation staff exchange scheme supported by Marie Sk\l{}odowska-Curie Actions at the Department of Electrical and Computer Engineering, New Jersey Institute of Technology, USA, in 2017. He was a Visiting Scholar with the School of Computer and Mathematics, Charles Sturt University, Australia, in 2018, and with the School of Information Technology, Deakin University, Australia, in 2019. He was a Research Fellow with the Department of Electronic Engineering, Tsinghua University, China, in 2020. His research interests include artificial intelligence, biometrics, cryptography, cybernetics, cybersecurity, forensics, information theory, mathematical optimisation, steganography, and watermarking.
\end{IEEEbiography}

\begin{IEEEbiography}[{\includegraphics[width=1in,height=1.25in,clip,keepaspectratio]{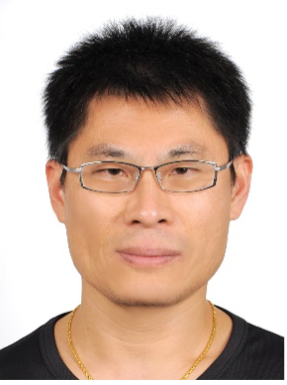}}]{Chun-Shien Lu}
received the Ph.D. degree in Electrical Engineering from National Cheng-Kung University, Tainan, Taiwan in 1998. He is a full research fellow (full professor) in the Institute of Information Science since March 2013 and the executive director in the Taiwan Information Security Center, Research Center for Information Technology Innovation, Academia Sinica, Taipei, Taiwan, since June 2024. His current research interests mainly focus on deep learning, AI security and privacy, and inverse problems. Dr. Lu serves as a Technical Committee member of Communications and Information Systems Security (CIS-TC) and Multimedia Communications Technical Committee (MMTC), IEEE Communications Society, since 2012 and 2017, respectively. Dr. Lu also serves as Area Chairs of ICASSP 2012--2014, ICIP 2013, ICME 2018, ICIP 2019--2025, ICML 2020, ICML 2023--2025, ICLR 2021--2026, NeurIPS 2022--2025, and ACM Multimedia 2022--2025, and Senior program committee of AAAI 2025--2026. Dr. Lu has owned four US patents, five ROC patents, and one Canadian patent in digital watermarking and graphic QR code. Dr. Lu won Ta-You Wu Memorial Award, National Science Council in 2007 and was a co-recipient of a National Invention and Creation Award in 2004. Dr. Lu was an associate editor of IEEE Trans. on Image Processing from 2010/12 to 2014 and from 2018/3 to 2023/6.
\end{IEEEbiography}

\begin{IEEEbiography}[{\includegraphics[width=1in,height=1.25in,clip,keepaspectratio]{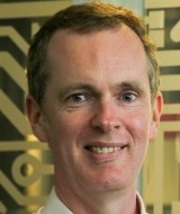}}]{Christopher Leckie} is currently a Professor with the School of Computing and Information Systems, The University of Melbourne. His research interests include artificial intelligence (AI), machine learning, anomaly detection, unsupervised learning, telecommunications, and cybersecurity. He has a strong interest in developing AI and machine learning techniques for a variety of applications in telecommunications, such as cyber security, network management, fault diagnosis, and the Internet of Things. He also has an interest in robust and scalable machine learning algorithms for problems, such as clustering and anomaly detection, with a focus on adversarial
machine learning and security analytics.
\end{IEEEbiography}

\begin{IEEEbiography}[{\includegraphics[width=1in,height=1.25in,clip,keepaspectratio]{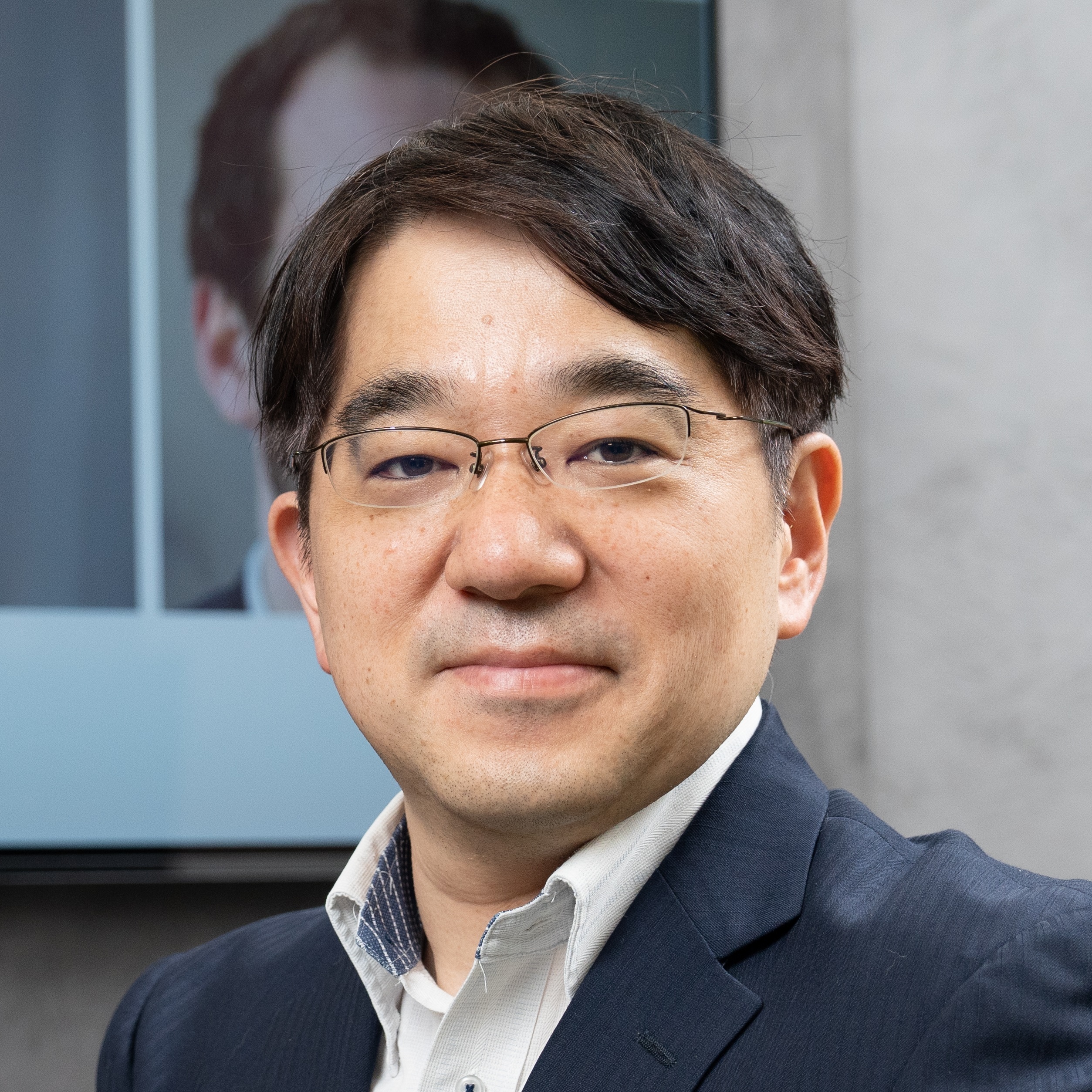}}]{Isao Echizen} (Senior Member, IEEE) received B.S., M.S., and D.E. degrees from the Tokyo Institute of Technology, Japan, in 1995, 1997, and 2003, respectively. He joined Hitachi, Ltd. in 1997 and until 2007 was a research engineer in the company's systems development laboratory. He is currently a director and a professor of the Information and Society Research Division, the National Institute of Informatics (NII), a director of the Global Research Center for Synthetic Media, the NII, a professor in the Department of Information and Communication Engineering, Graduate School of Information Science and Technology, the University of Tokyo, and a professor in the Graduate Institute for Advanced Studies, the Graduate University for Advanced Studies (SOKENDAI), Japan.  He was a visiting professor at the University of Freiburg, Germany, and at the University of Halle-Wittenberg, Germany. He is currently engaged in research on AI security, multimedia security, and multimedia forensics. He is a research director in the CREST FakeMedia project and in the K Program SYNTHETIQ X, Japan Science and Technology Agency (JST). He received the Commendation for Science and Technology by the Minister of Education, Culture, Sports, Science and Technology (Research category) in 2025, the Best Paper Award from the IEICE in 2023, the Best Paper Awards from the IPSJ in 2005 and 2014, the IPSJ Nagao Special Researcher Award in 2011, the DOCOMO Mobile Science Award in 2014, the Information Security Cultural Award in 2016, and the IEEE Workshop on Information Forensics and Security Best Paper Award in 2017. He was a member of the Information Forensics and Security Technical Committee of the IEEE Signal Processing Society. He is the IEICE Fellow, the IPSJ Fellow, the IEEE Senior Member, and the Japanese representative on IFIP and on IFIP TC11 (Security and Privacy Protection in Information Processing Systems), a vice president of APSIPA, and an editorial board member of the IEEE Transactions on Dependable and Secure Computing, the EURASIP Journal on Image and Video Processing, and the Journal of Information Security and Applications, Elsevier.
\end{IEEEbiography}
\appendices

\section{Evaluation Metrics}
\label{appx_metrics}
\subsection{Attack Success Rate (ASR)}
The ASR quantifies the effectiveness of an adversarial attack in altering the model's prediction. It is defined as the proportion of AEs that successfully cause the model to produce incorrect predictions. Mathematically, it can be expressed as:
\begin{equation}
    ASR=\frac{\text{Number of successful AEs}}{\text{Number of All AEs}}.
\end{equation}
A higher ASR indicates a more effective attack, as it reflects the ability of the AEs to mislead the model.

\subsection{Adversarial Detection Rate}
The adversarial detection rate refers to the proportion of AEs that are correctly identified by an adversarial detector. It is defined as:
\begin{equation}
    Detection\ Rate=\frac{\text{Number of AEs detected}}{\text{Number of All AEs}}.
\end{equation}
A lower detection rate indicates higher attack efficacy, as it reflects the ability of the AEs to evade detection.

\subsection{Structural Similarity Index (SSIM)}
The SSIM is a metric for measuring similarity between two images. It considers luminance, contrast, and structural information. The SSIM between two images $x$ (reference) and $y$ (distorted) is defined as:
\begin{equation}
    SSIM(x,y) = \frac{(2\mu_x\mu_y + C_1)(2\sigma_{xy} + C_2)}{(\mu_x^2 + \mu_y^2 + C_1)(\sigma_x^2 + \sigma_y^2 + C_2)},
\end{equation}
where:
\begin{itemize}
    \item \( \mu_x \) and \( \mu_y \): Mean intensity values of \( x \) and \( y \), respectively.
    \item \( \sigma_x^2 \) and \( \sigma_y^2 \): Variance of \( x \) and \( y \), respectively.
    \item \( \sigma_{xy} \): Covariance between \( x \) and \( y \).
    \item \( C_1 \) and \( C_2 \): Small constants added to stabilize the metric in the presence of weak denominators, defined as:
    \begin{equation}
        C_1 = (K_1 L)^2, \quad C_2 = (K_2 L)^2,
    \end{equation}
    where \( L \) is the dynamic range of the pixel values (e.g., 255 for 8-bit images), and \( K_1 \) and \( K_2 \) are small positive constants (commonly, \( K_1 = 0.01 \) and \( K_2 = 0.03 \)).
\end{itemize}

The SSIM value ranges from -1 to 1, where:
\begin{itemize}
    \item 1: Perfect similarity.
    \item 0: No similarity.
    \item Negative values indicate dissimilarity.
\end{itemize}

SSIM is typically calculated locally using a sliding window across the image, and the overall SSIM is the average of local SSIM values. Higher SSIM values denote subtler deviations from the reference images.

\subsection{Learned Perceptual Image Patch Similarity (LPIPS)}
The LPIPS is a perceptual similarity measure that evaluates the perceptual distance between two images. It uses deep neural network features to compare the visual content and correlates better with human perception than traditional metrics like SSIM.

The LPIPS distance between two images $x$ (reference) and $y$ (distorted) is computed as:
\begin{equation}
  \begin{aligned}
      &LPIPS(x,y)=\\
      &\sum_{l} \frac{1}{H_l W_l} \sum_{h=1}^{H_l} \sum_{w=1}^{W_l} w_l \| \hat{\phi}_l(x)_{h,w} - \hat{\phi}_l(y)_{h,w} \|_2^2,
  \end{aligned}  
\end{equation}
where:
\begin{itemize}
    \item \( l \): Layer index in the deep network.
    \item \( \phi_l(x) \): Feature map of \( x \) at layer \( l \) in the network.
    \item \( \hat{\phi}_l \): Normalized feature maps for each channel.
    \item \( H_l \), \( W_l \): Height and width of the feature map at layer \( l \).
    \item \( w_l \): Learned scalar weights for each layer \( l \) (trained to match human judgments).
\end{itemize}

LPIPS uses a pretrained deep neural network to extract feature maps from intermediate layers. The metric normalizes the feature maps and computes the difference between them, weighted according to the learned weights \( w_l \). The LPIPS value typically ranges from \( 0 \) (perfect perceptual similarity) to higher values indicating greater dissimilarity. LPIPS provides a perceptually grounded comparison, making it especially useful for evaluating subtle changes in visual quality.

\section{Experiment Settings}
The complete attack configurations are summarized in \Cref{tab_settings}, and all evaluated models are presented in \Cref{tab_arct}.

\begin{table}[!b]
    \centering
    \caption{Attack Settings.}
    \label{tab_settings}
    \begin{threeparttable}
        \setlength{\tabcolsep}{1.3mm}{\begin{tabular}{ccccc}
           \toprule
            Threat Model&Attack&Max. Q&$\epsilon$&Sparsity\\
            \midrule
            \multirow{3}{*}{\begin{tabular}{c}White-Box\\(Limited $\epsilon$)\end{tabular}}&APGD \cite{croce2020reliable}&\multirow{3}{*}{Unlimited}&\multirow{3}{*}{4/255}&\multirow{3}{*}{N/A}\\
            &AutoAttack \cite{croce2020reliable}&&&\\
            &GreedyPixel (Ours)&&&\\
            \midrule
            \multirow{5}{*}{\begin{tabular}{c}Black-Box\\(Limited $\epsilon$)\end{tabular}}&DeCoW \cite{lin2024boosting}&\multirow{5}{*}{20,000}&\multirow{5}{*}{4/255}&\multirow{5}{*}{N/A}\\
            &MCG \cite{yin2023generalizable}&&&\\
            &GFCS \cite{lord2022attacking}&&&\\
            &SQBA \cite{park2024hard}&&&\\
            &GreedyPixel (Ours)&&&\\
            \midrule
            \multirow{2}{*}{\begin{tabular}{c}Black-Box\\(Unlimited $\epsilon$)\end{tabular}}&BruSLe \cite{vo2024brusleattack}&\multirow{2}{*}{20,000}&\multirow{2}{*}{Unlimited}&1\%/0.1\%$^*$\\
            &GreedyPixel (Ours)&&&N/A\\
            \bottomrule
            \hline
        \end{tabular}}
        \begin{tablenotes}
            \item $^*$The sparsity for attacking CIFAR-10 and ImageNet was set to 1\% (10 pixels) and 0.1\% (50 pixels), respectively.
        \end{tablenotes}
    \end{threeparttable}
\end{table}

\begin{table}[!t]
    \centering
    \caption{Model architectures.}
    \label{tab_arct}
    \begin{threeparttable}
        \setlength{\tabcolsep}{2mm}{\begin{tabular}{cccc}
            \toprule
            Dataset&Architecture&Role&\begin{tabular}{c}Adversarial\\Training?\end{tabular}\\
            \midrule
            \multirow{7}{*}{CIFAR-10}&WideResNet-28-10 \cite{croce2021robustbench}&Target&NO\\
            &WideResNet-28-10 \cite{wang2023better}&Target&YES\\
            &WideResNet-82-8 \cite{bartoldson2024adversarial}&Target&YES\\
            &XCiT-S12 \cite{debenedetti2023light}&Target&YES\\
            &PreActResNet-18 \cite{wong2020fast}&Surrogate&YES\\
            &PreActResNet-18 \cite{gowal2021improving}$^*$&Surrogate&YES\\
            &RaWideResNet-70-16 \cite{peng2023robust}$^*$&Surrogate&YES\\
            \midrule
            \multirow{6}{*}{ImageNet}&ResNet-50 \cite{croce2021robustbench}&Target&NO\\
            &ViT-Base-Patch16 \cite{wu2020visual}&Target&NO\\
            &VGG19-BN \cite{simonyan2015very}&Target&NO\\
            &ConvNeXt \cite{singh2023revisiting}&Surrogate&YES\\
            &Swin \cite{liu2023comprehensive}$^*$&Surrogate&YES\\
            &ResNet-18 \cite{salman2020adversarially}$^*$&Surrogate&YES\\
            \bottomrule
        \end{tabular}}
        \begin{tablenotes}
            \item $^*$denotes the surrogate model was assessed for ablation studies only.
        \end{tablenotes}
    \end{threeparttable}
\end{table}

\section{Discussion}
\subsection{Actual Perturbation Values Applied to Effective Adversarial Examples}
Table~\ref{tab_ratio} summarizes our analysis of 1,000 adversarial examples produced by AutoAttack \cite{croce2020reliable}. We observe that 70.43\% of perturbation components under the $L_{\infty}$ threat model assume the extreme values $\pm\epsilon$, indicating that successful attacks frequently exploit the perturbation bounds. Motivated by this empirical finding, GreedyPixel constrains the search space to ${\pm\epsilon}$ per pixel and thereby reduces the combinatorial brute-force problem to a tractable pixel-wise greedy optimization.

\begin{table}[!t]
    \centering
    \caption{The ratio (\%) of perturbation values (in terms of their absolute values) in effective AEs @ $L_\infty$-norm $\epsilon=4/255$.}
    \label{tab_ratio}
    \begin{threeparttable}
        \setlength{\tabcolsep}{6.2mm}{\begin{tabular}{cccc}
            \toprule
            Value&CIFAR-10&ImageNet&Average\\
            \midrule
            4/255&\textbf{70.43}&\textbf{42.23}&\textbf{56.330}\\
            3/255&8.40&18.47&13.435\\
            2/255&6.46&13.59&10.025\\
            1/255&6.14&12.51&9.325\\
            0&8.57&13.2&10.885\\
            \bottomrule
        \end{tabular}}
        \begin{tablenotes}
            \item \textbf{Bold} denotes the largest proportion. 
        \end{tablenotes}
    \end{threeparttable}
\end{table}

\subsection{Manipulation on Sensitive Entities}
\begin{figure}[!t]
    \centering
    \includegraphics[width=\linewidth]{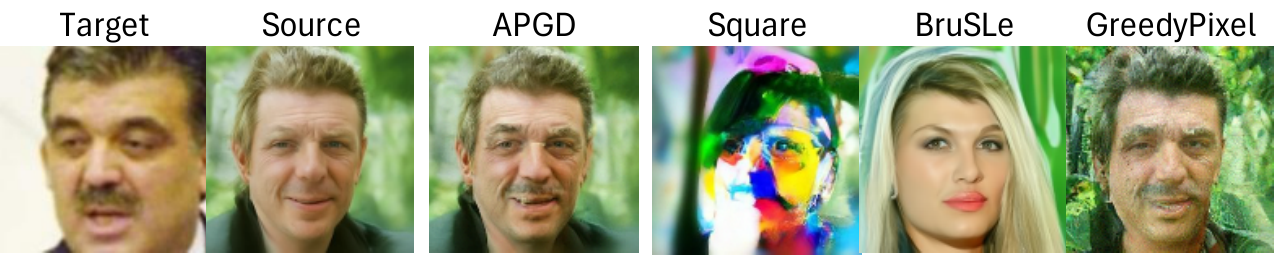}
    \caption{Model-inversion attack generations from the same source image to the same target image. Square \cite{andriushchenko2020square} and BruSLe \cite{vo2024brusleattack} attacks introduce significant differences.}
    \label{fig_blackMIA}
\end{figure}

Our recent work emphasizes the critical importance of fine-grained manipulations, like our GreedyPixel attack, particularly when targeting sensitive entities such as the latent codes in diffusion models. In contrast, other black-box methods, including Square \cite{andriushchenko2020square} and BruSLe \cite{vo2024brusleattack} attacks, often produce noticeable distortions or noise during the generation process, as illustrated in Figure~\ref{fig_blackMIA}.

\end{document}